\documentclass[a4paper]{article}
\usepackage{times}
\usepackage{latexsym}

\usepackage[english]{babel}
\usepackage[utf8x]{inputenc}
\usepackage{microtype}
\usepackage[colorinlistoftodos]{todonotes}
\usepackage[margin=1.2in]{geometry}

\usepackage{hyperref}
\usepackage{url}
\usepackage{mathtools}
\DeclarePairedDelimiter{\norm}{\lVert}{\rVert}
\newtheorem{theorem}{Theorem}[section]

\usepackage{amsmath, amssymb}
\usepackage{ntheorem}

\newtheorem{prop}{Proposition}
\newtheorem{proof}{Proof}
\usepackage{graphicx}
\usepackage{caption}
\usepackage{subcaption}
\usepackage{multirow}
\usepackage{color}
\newtheorem{conjecture}[theorem]{Conjecture}
\usepackage{soul}

\usepackage{natbib}

\title{A Meta-Learning Perspective on Transformers for Causal Language Modeling}

\author{Xinbo Wu \\
  University of Illinois Urbana-Champaign  \\
  \texttt{xinbowu2@illinois.edu} \\\and
  Lav R.\ Varshney \\
  University of Illinois Urbana-Champaign \\
  \texttt{varshney@illinois.edu} \\}
\date{}
\begin{document}
\maketitle
\begin{abstract}
 The Transformer architecture has become prominent in developing large causal language models. However,  mechanisms to explain its capabilities are not well understood. Focused on the training process, here we establish a meta-learning view of the Transformer architecture when trained for the causal language modeling task, by explicating an inner optimization process within the Transformer. Further, within the inner optimization, we discover and theoretically analyze a special characteristic of the norms of learned token representations within Transformer-based causal language models. Our analysis is supported by experiments in various settings. 
\end{abstract}

\section{Introduction}
The Transformer architecture for neural networks is based on a self-attention mechanism \citep{Transformer} and has been widely used in natural language processing, computer vision, and scientific discovery \citep{BERT, GPT-2, VIT, BERT_protein}, among other topical areas. Causal language modeling (CLM) aims to predict the next element in a sequence in an autoregressive manner and is one of the most important applications of the Transformer model. Large language models (LLMs) based on the Transformer architecture and CLM have shown impressive capabilities in many fields that have been considered difficult challenges in artificial intelligence (AI) \citep{ChatGPT,GPT-4}. However, the underlying mechanisms of Transformer-based causal language models are still not well understood.

Meta-learning seeks to learn how to learn by leveraging common knowledge across various tasks, often through a bi-level optimization~\citep{metaLSTM, MAML, FranceschiFSGP2018}. Within this framework, an inner process mirrors a typical model optimization process such as stochastic gradient descent (SGD) \citep{SGD}, whereas an outer process is employed to learn hyperparameters for the inner process. Studying Transformer models from a meta-learning perspective is an emerging research direction. Some recent works of \citet{in_context_gd, in_context_preconditioned_gd, in_context_mesa, in_context_gpt} have attempted to interpret in-context learning in Transformer-based language models from a meta-learning perspective by identifying a bi-level optimization process within in-context learning. However, they rely either on simplifications of the original Transformer model such as linear attention, or on special constructions of learned parameters. We make very limited assumptions to analyze a model that is as close to the Transformer used in practice as possible.

More importantly, rather than focusing on in-context learning that is regarded as an emergent capability of Transformer-based CLM models, we explore whether we can discover a meta-learning process within the \emph{training} of a Transformer-based CLM model. We attempt to give a plausible bi-level optimization process for a Transformer model trained for the CLM task. Especially, we find there is an inner optimization process within the forward pass of the model, which optimizes for both clustering and the CLM objective. To investigate this inner optimization process from the meta-learning perspective, we visualize the evolution of the token representations within Transformer-based CLM models. We uncover a characteristic of the evolution of the norms of token representations learned in Transformer-based CLMs, which may indicate a special optimization trajectory.

We summarize our contributions as follows. 
\begin{itemize}
  \item We derive a plausible inner optimization process for clustering and CLM in a Transformer model trained for the CLM task, which is also supported by comprehensive evidence from our experiments in various settings.
  \item Building on the inner optimization process, we present a meta-learning view of a Transformer-based CLM model.
  \item Through experimental investigation of the inner optimization process, we discover a special characteristic of norms of the token representations learned by the Transformer-based CLM model, which may indicate a special optimization trajectory. We analyze this characteristic and further investigate it through experiments on real-world LLMs and datasets.
\end{itemize}

\section{Preliminaries}\label{sec:preliminary}
Let us first formulate a linear projection weight matrix in a Transformer model by expanding it via its gradient descent learning dynamics. A weight matrix ${W\in \mathbb{R}^{d_{out}\times d_{in}}}$ in a Transformer model trained by standard stochastic gradient descent using ${T}$ training inputs $x_{1},\ldots,x_{T}$ to optimize a loss function ${L}$ via backpropagation can be represented as in~\citet{duality_nn}:
\begin{equation}
    W = W_{0} - \sum_{t=1}^{T}\eta_{t}\nabla_{y_{t}}L\otimes x_{t}
\label{eqt:W_GD_1}
\end{equation}
where ${W_{0}\in \mathbb{R}^{d_{out}\times d_{in}}}$ is the weight initialization, ${\eta_{t} \in \mathbb{R}}$ is the learning rate, ${y_{t} = W_{t}x_{t}}$ is the output of the linear transformation of ${x_t}$ via ${W_{t}}$ at step ${t}$ and ${\nabla_{y_{t}}L\otimes x_{t} = \nabla_{y_{t}}Lx_{t}^\mathsf{T}}$ is the outer product between ${\nabla_{y_{t}}L}$ and ${x_{t}}$. We treat ${-\eta_{t}\nabla_{y_{t}}L\otimes x_{t}}$ as an error signal. Note that with our expression, training on one single token corresponds to one training step. 

For simplicity, we assume ${W_{0}}$ is very small such that we can use a zero matrix to approximate it. Then, we have: 
\begin{equation}
    W \approx - \sum_{t=1}^{T}\eta_{t}\nabla_{y_{t}}L\otimes x_{t}.
\label{eqt:W_GD_approx}
\end{equation}

The assumption of zero initialization may be better approximated by larger models. It is common practice to initialize weights with smaller values for larger dimensions to control the initial variance of the output, thereby fostering a stable learning process and mitigating problems like exploding gradients. Numerous initialization methods commonly employed, such as Xavier~\citep{Xavier} and Kaiming~\citep{Kaiming_init}, incorporate a scaling factor proportional to the inverse of the number of dimensions in either or both the input and output. 

Next, we analyze various components of a Transformer-based CLM model. \citet{Transformer} provide a more detailed introduction to the Transformer model. 

\section{Analysis of a Language Modeling Head}\label{sec:language_head_knn}
A language modeling (LM) head is the final layer of a Transformer-based CLM model and is responsible for converting its hidden token representation into a distribution over its vocabulary. The LM head is implemented by a linear layer. We can open it via \eqref{eqt:W_GD_approx} as follows:
\begin{equation}
    W_{LH} \approx - \sum_{t=1}^{T}\eta_{t}^{L+1}\nabla_{z_{t}^{L+2}}L\otimes z_{t}^{L+1}.
\label{eqt:W_GD_approx_lh_1}
\end{equation}
where $z_{t}^{L+1}$ is the output of the $L$th Transformer layer at step $t$ by assuming we have $L$ Transformer layers. Here, $\nabla_{z_{t}^{L+2}}L$ is the gradient of the cross-entropy loss with respect to the logits, $z_{t}^{L+2}$. Note that $z_{t}^{L+2}$ is the projected output and will go into a softmax function for sampling the next token.

By assuming the next token label at step $t$ used to compute the loss is a one-hot vector $y_{t}$, the $s_{t}$ is the softmax output vector and expanding $\nabla_{z_{t}^{L+2}}L$, we get:
\begin{equation}
    z_{T+1}^{L+2} = W_{LH}z^{L+1} \approx \left[\sum_{t=1}^{T}\eta_{t}^{L+1}(y_{t}-s_{t}) (z_{t}^{L+1})^\mathsf{T}\right]z_{T+1}^{L+1}.
\label{eqt:W_GD_approx_lh_2}
\end{equation}

Suppose our target is $y_{T+1}$. To make a good prediction, the dimension in the current logit vector, $z_{T+1}^{L+2}$ corresponding to the current label has to have the highest value among other all dimensions. Please note the softmax outputs are non-negative. Therefore, we expect the weights computed by the dot product, $(z_{t}^{L+1})^\mathsf{T}z_{T+1}^{L+1}$ to be large enough for those $y_{t}$ being the same as $y_{T+1}$ so the overall sum produces the desired logits. This process essentially approximates a K-nearest neighbor algorithm (KNN), in which distances between training examples in the history and current example are computed by the dot product, and historical one-hot labels are aggregated by using their respective distances to generate an inference for the current example. 

This also indicates that a model should organize the token representation, $z_{T+1}^{L+1}$ close to at least some examples with the same label in the training history in the sense of high dot product scores, so we get logits favoring a correct prediction. That is, these examples form a cluster within the hidden representation space. Further, note that a small or even negative dot product results from examples with different labels and will contribute less to or reduce the values in their corresponding dimensions of the logit vector, which may also lead to a good prediction. 

Ideally, a properly learned LM head produces similar logits corresponding to the same label such that they could potentially be clustered together. This means the LM head performs clustering in the sense that the projected inputs by the LM head are clustered based on their respective labels.

\citet{clustering_llm} recently showed a clustering phenomenon within the hidden states of Transformer-based CLMs trained for an instruction-following task. Figure~\ref{fig:learning_dynamics_f1_val} also shows a similar clustering phenomenon for a regular CLM task; experimental details are in Section~\ref{sec:experiment_sythetic}. These findings along with our understanding of the LM head from a KNN perspective inspire us to wonder whether a Transformer-based CLM model performs clustering iteratively through its layers such that we can get expected hidden representations for the approximate KNN to work. 

\section{Analysis of a Transformer Layer} \label{sec:Transformer_gc}

In this section, we analyze a Transformer-based CLM model from its mathematical formulations to seek mechanisms leading to the clustering phenomenon. The model consists of multiple Transformer layers and each Transformer layer is formed by connecting a multi-head self-attention (MHSA) sublayer and a feedforward network (FFN) sublayer. A detailed investigation of them will follow. 

\subsection{Analysis of a Multi-head Self-attention Layer}
Next, we assume the number of dimensions of a token representation within a Transformer model to be ${d_{model}}$. By following \citet{Transformer}, we write and expand the formulation of the multi-head attention module for the CLM task with ${z^{l} \in \mathbb{R}^{n\times d_{model}}}$ being the representations of current ${n}$ sequential tokens at layer ${l}$: 
\begin{equation}\label{eqt:MHSA_1}
    \begin{aligned}
       \text{MHSA}^{l}(z_{1:n}^{l}) = \sum_{h=1}^{H}W_{O}^{lh}\sum_{i=1}^{n}v_{i}^{lh}\text{softmax}((k^{lh})^\mathsf{T}q_{n}^{lh})_{i}
    \end{aligned}
\end{equation}
where ${\text{MHSA}^{l}}$ is the parameterized multi-head self-attention module at layer ${l}$. Here, ${W_{O}^{lh}} \in \mathbb{R}^{d_{model}\times d_{head}}$, ${W_{V}^{lh}}$, ${W_{K}^{lh}}$, and ${W_{Q}^{lh} \in \mathbb{R}^{d_{head}\times d_{model}}}$ are output, value, key, and query projection matrices of head ${h}$ at layer ${l}$ respectively. We have ${d_{model} = Hd_{head}}$ and ${H}$ as the total number of heads. The $i$th value ${v_{i}^{lh}}$, key ${k_{i}^{lh}}$, and query ${q_{i}^{lh}}$ of head ${h}$ at layer ${l}$ are computed by ${v_{i}^{lh} = W_{V}^{lh}z_{i}^{l}}$, ${k_{i}^{lh} = W_{K}^{lh}z_{i}^{l}}$ and ${q_{i}^{lh} = W_{Q}^{lh}z_{i}^{l}}$ respectively. Further, ${k^{lh} = \left[k_{1}^{lh}, \ldots, k_{n}^{lh}\right]}$ is the key matrix of head ${h}$ at layer ${l}$. The $i$th element of the \text{softmax} output is denoted ${\text{softmax}((k^{lh})^\mathsf{T}q_{n}^{lh})_{i}}$. Throughout, we ignore the scalar factor within the \text{softmax} attention and any bias terms, to simplify analysis.

We can arrive at the following approximation by derivation shown in Appendix~\ref{appx:MHSA}:
\begin{equation}
\begin{aligned}
    \text{MHSA}^{l}(z_{1:n}^{l}) \approx  -\sum_{t=1}^{T}\eta_{t}^{l}\sum_{h=1}^{H}w_{t}^{lh}A_{t}^{lh}\nabla_{\hat{z}_{t}^{l}}L 
\label{eqt:MHSA_approx_2}
\end{aligned}
\end{equation}
where ${w_{t}^{lh} = (\hat{v}_{t}^{lh})^\mathsf{T}\sum_{i=1}^{n}v_{i}^{lh}\text{softmax}((k^{lh})^\mathsf{T}q_{n}^{lh})_{i}}$ is a weighting factor and ${A_{t}^{lh} = (\frac{\partial \hat{z}_{t}^{l}}{\partial \hat{y}_{t}^{lh}})^\mathsf{T} \in \mathbb{R}^{d_{model}\times d_{model}}}$ is a transformation matrix. We denote a historical vector at a step in history with a hat ${\hat{\cdot}}$. ${L}$ is assumed to be the CLM loss throughout and ${\hat{v}_{t}^{lh}}$ refers to a historical value of head ${h}$ and layer ${l}$ at step ${t}$ in the training history.  Note that ${t}$ should not be confused with a token position within a sequence. As before, ${\hat{y}_{t}^{lh}} = W_{O,t}^{lh}\hat{v}_{t}^{lh} \in \mathbb{R}^{d_{model}}$ and ${\eta_{t}^{lh}}$ is the learning rate of the corresponding weight matrix ${W_{O,t}^{lh}}$ of head ${h}$ and layer ${l}$ at step ${t}$. 

We can interpret the formulation in \eqref{eqt:MHSA_approx_2} as a query-key retrieval of a value via the softmax self-attention and then, using the retrieved value as a query and the historical values as keys to obtain a weighted average of transformed historical gradients with respect to the historical inputs to the module. This formulation is similar to what we found in~\eqref{eqt:W_GD_approx_lh_2}. In principle, the weighted average of transformed gradients could implement an approximation of the transformed gradient with respect to the current input ${z_{n}^{l}}$ by its neighborhood in a soft manner with closeness measured by the weighting factor. A gradient with respect to an input essentially provides an updating direction for the input to minimize a certain loss. This formulation is essentially similar to the one we found in~\eqref{eqt:W_GD_approx_lh_2} and intuitively the \text{MHSA} module could perform clustering using a similar mechanism through the neighborhood of its input in the history. 

Another intuition is that if a representation of every example belonging to a specific cluster is computed based on a similar neighborhood, these representations are likely to be similar and easily clustered. Further, this multi-retrieval process is conducted for different heads and the final gradient is also aggregated over different heads. Many methods treat a transformed gradient as a pre-conditioned gradient that captures curvature information~\citep{meta_curvature, fisher_nn, fisher_conv}, such as Newton's method for second-order optimization~\citep{numerical_opt}.

A formulation of a MHSA sublayer is shown in \eqref{eqt:MHSA_layer_1}. We can represent the $n$th token representation at the $l$th layer after the MHSA module with residual connection as follows:
\begin{equation}
\begin{aligned}
   z_{n}^{l+\frac{1}{2}} = Norm(z_{n}^{l} + \text{MHSA}^{l}(z_{1:n}^{l})) 
   \approx Norm(z_{n}^{l} -\sum_{t=1}^{T}\eta_{t}^{l}\sum_{h=1}^{H}w_{t}^{lh}A_{t}^{lh}\nabla_{\hat{z}_{t}^{l}}L )
\label{eqt:MHSA_layer_1}
\end{aligned}
\end{equation}
where ${Norm(\cdot)}$ is a normalization function such as LayerNorm \citep{layernorm} or RMSNorm \citep{RMSNorm}. We omit its parameter differences across layers to not abuse notations.

From \eqref{eqt:MHSA_layer_1}, we can see an approximate transformed gradient is used to update the current token representation. From our analysis of the LM head in Section \ref{sec:language_head_knn}, better clustering of examples may indicate a lower CLM loss. Therefore, we hypothesize that the gradients could direct the token representations to form clusters to minimize the loss. We will provide empirical evidence via experiments to support this interpretation.    

\subsection{Analysis of a Feedforward Network}
We can perform a similar analysis to the feed-forward network sublayer as follows and see the full details of the derivation in Appendix~\ref{appx:FFN}:
\begin{equation}\label{eqt:FFN_1}
    \begin{aligned}   
        \text{FFN}^{l}(z_{n}^{l+\frac{1}{2}}) = W_{2}^{l}\phi(W_{1}^{l}z_{n}^{l+\frac{1}{2}})
        \approx 
        -\left[\sum_{t=1}^{T}(\frac{\partial \hat{z}_{n}^{l+\frac{1}{2}}}{\partial \hat{b}_{t}^{l}})^\mathsf{T}\eta_{t}^{l+\frac{1}{2}}\nabla_{\hat{z}_{t}^{l+\frac{1}{2}}}L\otimes \hat{a}_{t}^{l}\right]a_{n}^{l} \\\rightarrow  
        z_{n}^{l+1} = Norm(z_{n}^{l+\frac{1}{2}} + \text{FFN}^{l}(z_{n}^{l+\frac{1}{2}}))\\ \approx = 
        Norm(z_{n}^{l+\frac{1}{2}} - \sum_{t=1}^{T}\eta_{t}^{l+\frac{1}{2}}w_{t}^{l+\frac{1}{2}}B_{t}^{l}\nabla_{\hat{z}_{t}^{l+\frac{1}{2}}}L) 
    \\= Norm(z_{n}^{l+\frac{1}{2}} - \Bar{B^{l}\nabla_{z_{n}^{l+\frac{1}{2}}}L})
    \end{aligned}
\end{equation}
where ${W_{2}^{l} \in \mathbb{R}^{d_{model}\times d_{ff}}}$ and ${W_{1}^{l} \in \mathbb{R}^{d_{ff}\times d_{model}}}$ are weight matrices of the first and second layer in the FFN module at Transformer layer ${l}$ and ${\phi}$ is a non-linear activation function. Some popular choices of nonlinear activation functions are ReLU~\citep{ReLU} and SwiGLU~\citep{SwiGLU}. Here,
${a_{n}^{l} = \phi(W_{1}^{l}z_{n}^{l+\frac{1}{2}})}$ and ${L}$ is the CLM loss, ${\hat{a}_{t}^{l}}$ is a historical output of the first layer at Transformer layer ${l}$ at step ${t}$. Further, ${\hat{b}_{t}^{l} = W_{2, t}^{l}\hat{a}_{t}^{l}} \in \mathbb{R}^{d_{model}}$, ${\eta_{t}^{l+\frac{1}{2}}}$ is the corresponding learning rate at step ${t}$, ${w_{t}^{l+\frac{1}{2}} = (\hat{a}_{t}^{l})^\mathsf{T}a_{n}^{l}}$ is a weighting factor, and ${B_{t}^{l} = (\frac{\partial \hat{z}_{t}^{l+\frac{1}{2}}}{\partial \hat{b}_{t}^{l}})^\mathsf{T} \in \mathbb{R}^{d_{model}\times d_{model}}}$ is a transformation matrix and ${\Bar{B^{l}\nabla_{z_{n}^{l+\frac{1}{2}}}L}}$ is an approximation of the transformed gradient with respect to the current input ${z_{n}^{l+\frac{1}{2}}}$.

Similar to the previous analysis, we can interpret the \text{FFN} module as a query-key retrieval of a weighted average of transformed historical gradients with respect to the historical module input. The \text{FFN} module with the residual connection could approximate a transformed gradient update to move the current token representation toward a cluster that optimizes the CLM loss ${L}$. 

Overall from its mathematical formulations, a Transformer layer can be understood as approximately performing two types of transformed gradient updates via the \text{MHSA} and the \text{FFN} modules, respectively. Particularly, the token representations are normalized after each type of transformed gradient update. Thus, based on this understanding, the forward pass of a series of Transformer layers trained for the CLM task can be viewed as an inner optimization process by iteratively moving the current token representation toward a cluster, which helps to minimize the CLM loss ${\bar{L}}$ at the same time. 

In this section, we have shown the plausibility of an inner optimization process for both clustering and minimizing the CLM loss within a Transformer trained for the CLM task from its mathematical formulations. In Sections \ref{sec:experiment_sythetic} and \ref{sec:experiment_realistic}, we will present empirical evidence to support this understanding. 

\begin{figure*}[!htb]
\centering
\includegraphics[width=\linewidth]{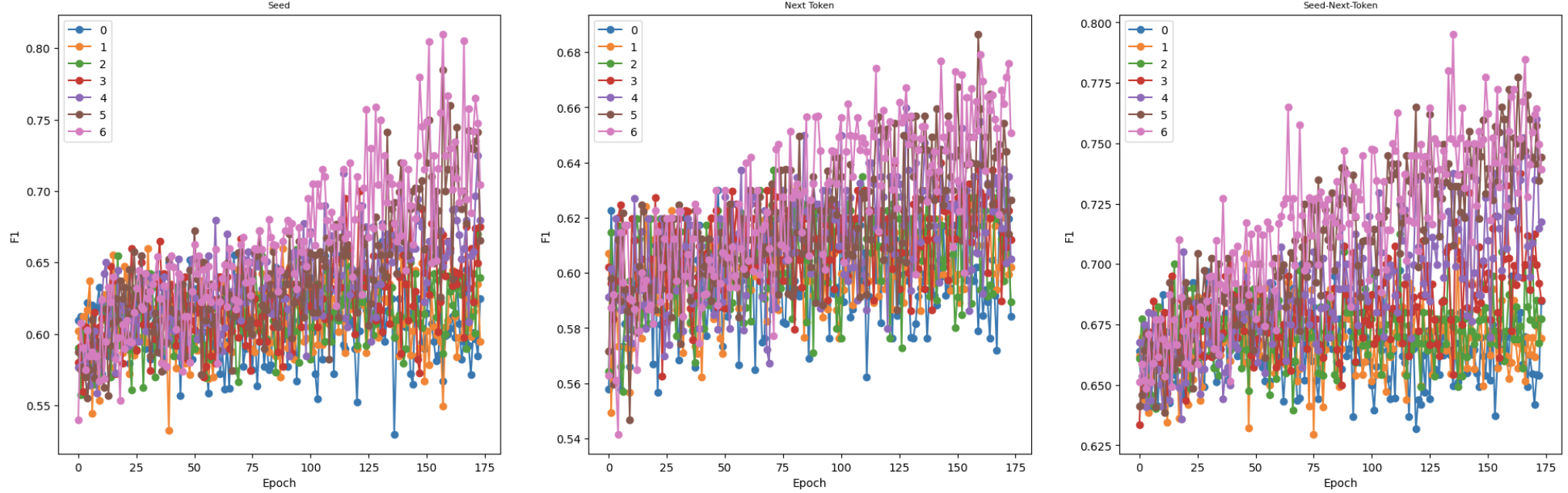}
\caption{Clustering analysis on the validation set across different layers throughout the training process. Different columns indicate different ground truth labels: seed, next token, and their combination (Seed-Next-Token). The legend shows layers. We only present results on F1 score here and refer to results on other evaluation metrics and other data split in Figure~\ref{fig:learning_dynamics_all} in the Appendix. Each dot illustrates a data point.}
\label{fig:learning_dynamics_f1_val}
\end{figure*}

\section{Experiments with Synthetic Language Data}\label{sec:experiment_sythetic}
Studying the mechanisms proposed in Section~\ref{sec:Transformer_gc} involves recording and computations based on a long training history. It is infeasible to use large-scale real-world language data due to excessive memory consumption, so we opt to use a small synthetic language dataset, which allows efficient capture of its training history. We construct a synthetic language dataset by using regular expressions inspired by \citet{clustering_llm}. Different from their simplified instruction-following task, we focus on a language modeling task. 

To build up a language dataset, we sample a textual sequence based on a sampled regular expression, where a regular expression can be considered as a simple grammar rule. We randomly construct 10 regular expressions as seeds for further sequence sampling. The resulting dataset is split into a training set and a validation set. More details about the data construction and its statistics are in Appendix~\ref{sec:appendix:synthetic_data_construction}. We train a six-layer Transformer model following the GPT-2 design \citep{GPT-2} on this dataset for the CLM task. To best mimic the learning dynamics described in Section~\ref{sec:preliminary}, we use a plain SGD optimizer with a constant learning rate of 1.0 and no weight decay regularization. More implementation details are in Appendix~\ref{sec:appendix:hyperparameters}. 

Once the training is done, following \citet{clustering_llm}, we perform a clustering analysis by using the KMeans algorithm from the scikit-learn package \citep{sklearn} on the token representation of the last tokens within many sequences, since the last token sees the whole sequence and has enough contexts. We measure clustering performances based on three possible ground truths: the next token label, the seed label, and a combination of them. We use F1, adjusted Rand index (ARI), and adjusted mutual information (AMI) as metrics to evaluate clustering performance.

From Figure~\ref{fig:learning_dynamics_f1_val}, we can see that there exist clustering phenomena under three different ground truths. Cluster performance generally improves across layers and training epochs. These results occur on both the training and validation data and support our view of the forward process of a Transformer-based CLM model as an iterative clustering process and this process is gradually optimized through the training process. Surprisingly, we also found clusters of examples generated from the same seed, in addition to our expected clusters based on the next token labels.  \citet{clustering_llm} found a similar result from their study of Transformer-based CLM models on a simplified instruction-following setting. This may reveal an inductive bias of Transformer-based CLM models. 

In addition, we observe the model achieves better clustering performance based on a combination of the seed label and next token label than that based on only the next token label, which may suggest that the model tends to group examples based on the combination label instead of solely the next token label. This gives us a picture of the cluster structure: the examples are clustered according to the seed labels and there exist clusters based on next token labels within each seed-specific cluster. 

From our analysis in Section~\ref{sec:language_head_knn}, clustering examples with the same next token label directly benefit the next token prediction objective. Therefore, we hypothesize that the gradual improvements in clustering performance across layers could lead to improvements of the CLM loss across layers. To verify this hypothesis, we feed the current token representation of each layer to the linear LM head to make the next token prediction and measure the corresponding cross-entropy loss via using the next token label as follows:
\begin{equation}
    \begin{aligned}
        &L_{n}^{l} = \text{CrossEntropy}(\text{softmax}(LH(z_{n}^{l})),y_{n})
    \label{eqt:inner_loss}
    \end{aligned}
\end{equation}\label{inner_loss_equation}
where $z_{n}^{l}$ is the representation of the $n$-th (current) token at layer $l$, $LH$ is the linear LM head and $y_{n}$ is the true label of the current token.

\begin{figure*}[!htb]
\centering
\begin{subfigure}{0.45\textwidth}
  \centering
  \includegraphics[width=0.8\linewidth]{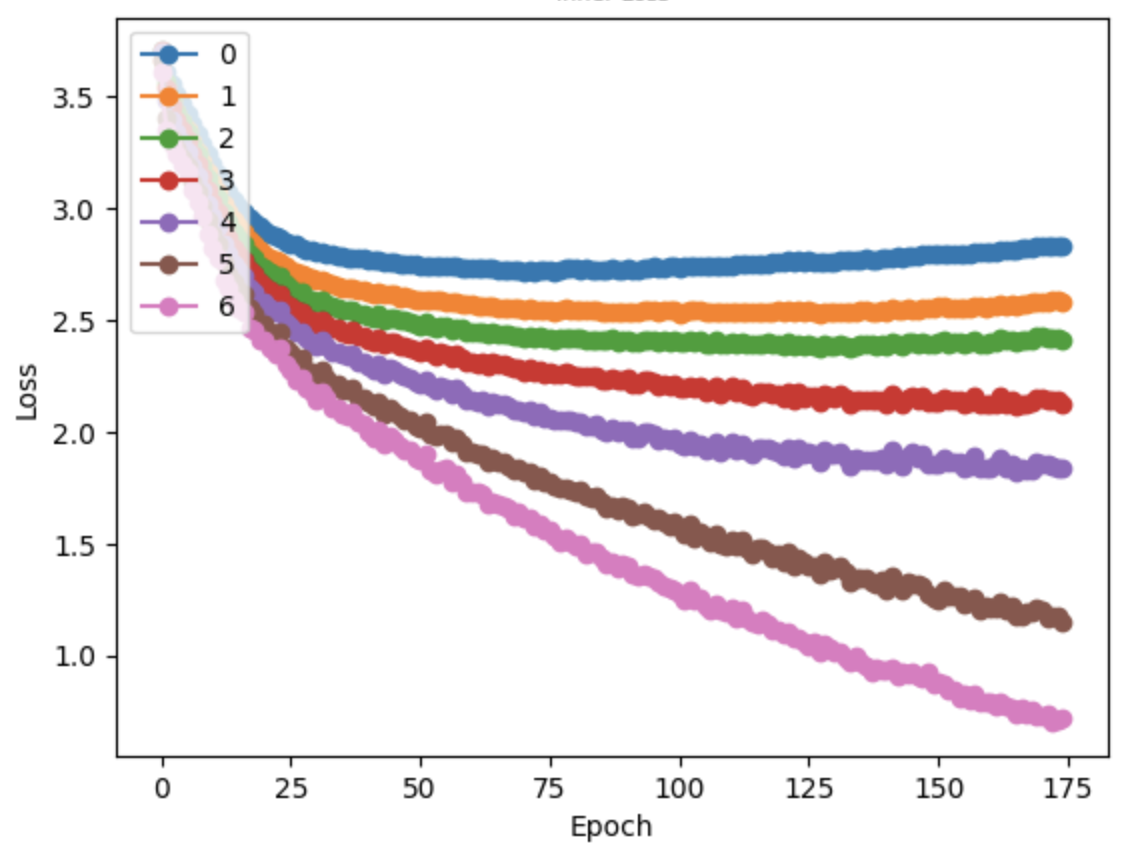}
  \caption{Inner Loss}
  \label{fig:inner_loss_SGD}
\end{subfigure}%
\begin{subfigure}{0.5\textwidth}
  \centering
  \includegraphics[width=\linewidth]{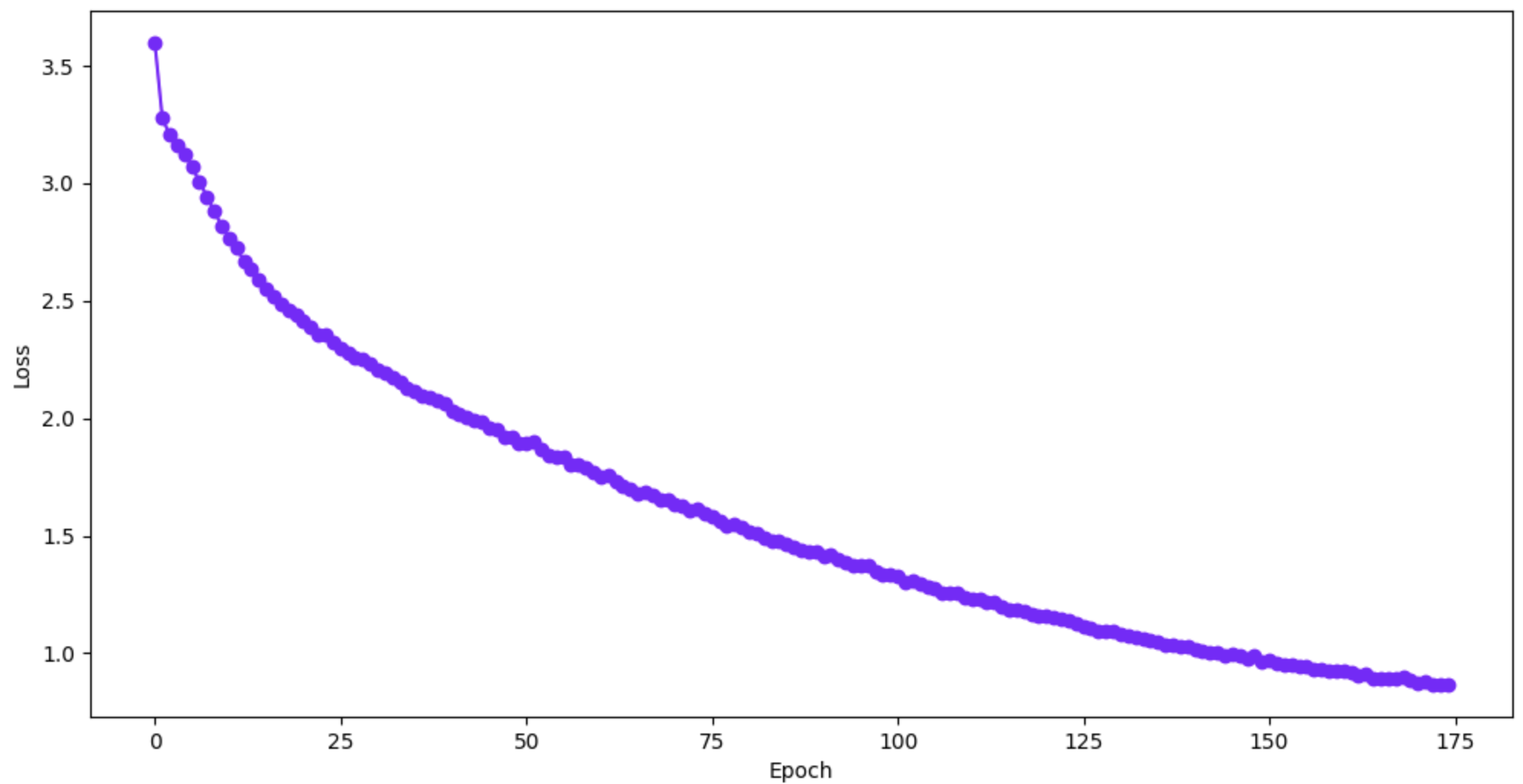}
  \caption{Outer Loss (Training Loss)}
  \label{fig:train_loss_sgd_train}
\end{subfigure}%
\caption{Bi-level Optimization process within a Transformer-based CLM model. (a) shows inner optimization losses across layers and training epochs computed according to \eqref{inner_loss_equation} and aggregated from training examples. (b) illustrates the losses of the outer optimization process throughout the training process, which explicitly optimizes for the CLM task. The outer loss is identical to the training loss of the model.}
\label{fig:bi-level_optimization_SGD_figure}
\end{figure*}

The results in Figure~\ref{fig:train_loss_sgd_train} show the inner CLM loss indeed decreases across layers, which is consistent with our hypothesis of the forward process as an inner optimization process for the CLM loss. 


\begin{figure*}[!htb]
\centering
\includegraphics[width=\linewidth]{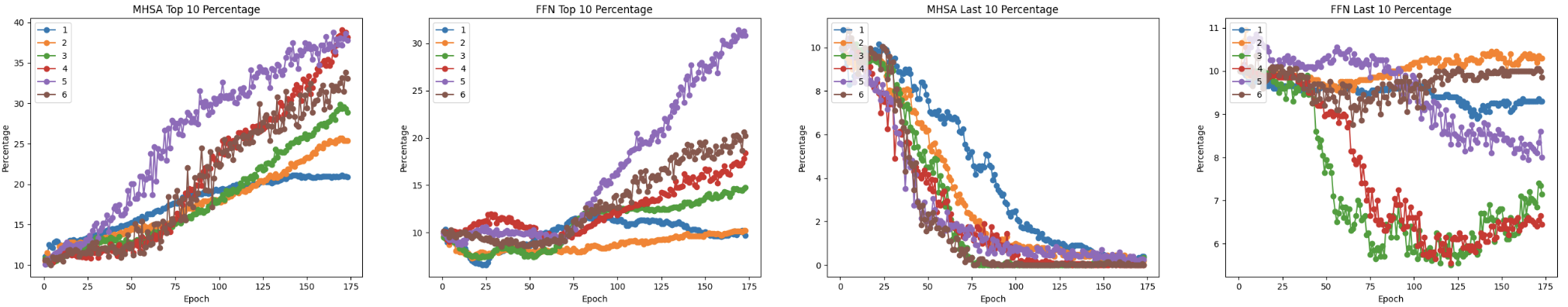}
\caption{Attention over the history throughout the training evaluated using seed identity as ground truth: we compute the attentions from an unseen instance (the last token of each sequence) from the validation set to all of the historical instances and calculate the percentage of the top 10  and last 10 attended instances having the same label. The final measurement is aggregated across different heads, token positions, and instances and is reported per layer. See the same analysis based on other ground truths in Appendix~\ref{fig:att_lout=1}}
\label{fig:att_lout=0}
\end{figure*}

The mathematical analysis presented in Section ~\ref{sec:Transformer_gc} outlines possible mechanisms to perform the clustering by the model. We verify these mechanisms by investigating relevant components of the model. The weighting factors in \eqref{eqt:MHSA_approx_2} and \eqref{eqt:FFN_1} are our special focus and can be viewed as attention scores over the historical examples. We study the Transformer components including the \text{MHSA} and the \text{FFN} modules. For each component, we compute the top 10 attended training examples in history by an unseen example according to its weighting factor and measure the percentage of them belonging to the same ground truth cluster label as the unseen example. We call this measurement the same label percentage. The same measurements are performed on the last 10 attended examples by following our observation in Section~\ref{sec:language_head_knn}. We consider the seed labels and the combination labels instead of the next token labels since their better clustering performances may indicate more accurate cluster structures. Please note that the baseline value is 10 percent when using the seed label because there are 10 different seeds. Similarly, the setting with the combination label has a baseline of around 2 percent. 

Figure~\ref{fig:att_lout=0} demonstrates that for the top 10s, the same label percentage is larger than its baseline and increases along the training process in most layers for both the \text{MHSA} and the \text{FFN} modules and under all of the settings. An opposite trend occurs for the last 10 attended examples as we expect, which shows the least attended examples are examples with different cluster labels. These results confirm our understanding of the mechanism used by the model to perform clustering. 

\section{Experiments in a Realistic Setting}\label{sec:experiment_realistic}
We have seen evidence of our understanding of the hidden mechanism of a Transformer-based CLM model proposed in Section~\ref{sec:Transformer_gc} on a small-scale model trained on synthetic language data. Next, we seek evidence of our view of the forward process of a Transformer-based CLM model as an inner optimization process in a realistic setting to supplement our results based on synthetic language data. Please note we do not expect our findings on small-scale models to completely transfer to the large-scale models trained on massive complicated language data. Some works \citep{llm_differently, why_llm_differently} have shown large models behave differently from small models. Also, the Transformer-based CLM model is highly intricate, so we do not aim to give a full picture of the hidden mechanisms used by the model but at least part of it. 

Our experiments are based on three popular Transformer-based causal language models: GPT-2~\citep{GPT-2} (released under a modified MIT license), LLaMa-7B, and LLaMa-13B~\citep{Llama} (both released under a Meta license). Their parameter counts are 1.5 billion (GPT-2), 7 billion (LLaMa-7B), and 13 billion (LLaMa-13B) respectively. We use test sets of Wikitext-103~\citep{wikitext-103} and 1 Billion Words (1BW)~\citep{1BW}, two common language benchmark datasets in our experiments. The wikitext-103 and 1BW datasets were constructed from Wikipedia articles and news crawl data respectively, so they follow different data distributions. To have enough context for the next token prediction task, we only keep data with lengths more than or equal to 10 words. Due to limitations on computational resources, we only randomly chose 5,000 samples from the test set of 1BW. 


\begin{figure*}
\centering
\begin{subfigure}{.3\textwidth}\label{fig:loss_GPT2_wikitext-103}
  \centering
  \includegraphics[width=0.75\linewidth]{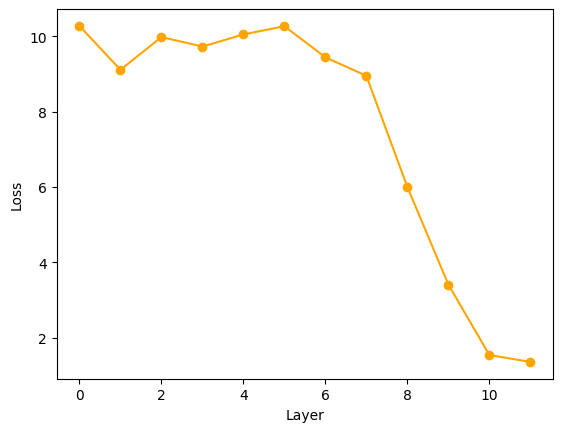}
  \caption{GPT-2}
\end{subfigure}%
\begin{subfigure}{.3\textwidth}\label{fig:loss_llama_7B_wikitext-103}
  \centering
  \includegraphics[width=0.75\linewidth]{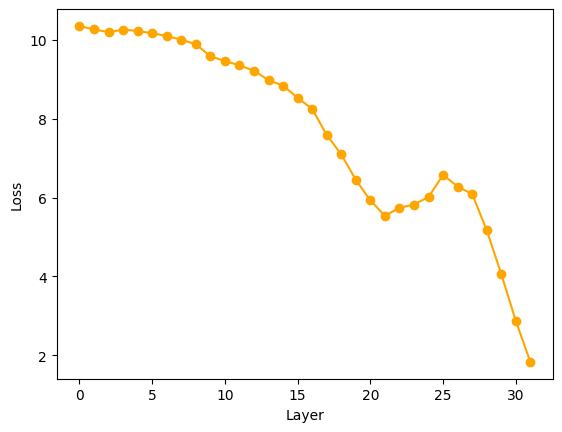}
  \caption{LLaMa-7B}
\end{subfigure}
\begin{subfigure}{.3\textwidth}\label{fig:loss_llama_13B_wikitext-103}
  \centering
  \includegraphics[width=0.75\linewidth]{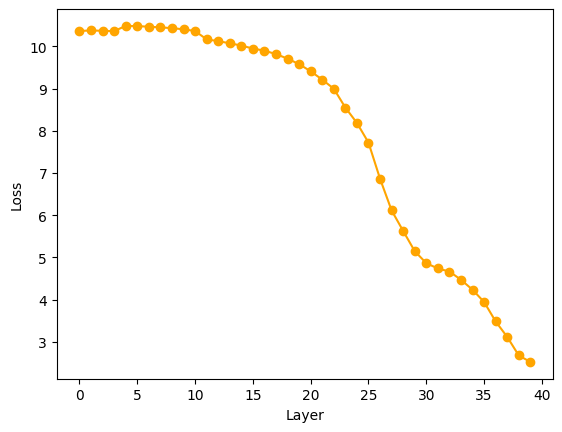}
  \caption{LLaMa-13B}
\end{subfigure}
\caption{Inner Optimization loss across layers based on The Wikitext-103 Dataset: the x-axis is layer numbers and y-axis is the inner loss averaged over samples. The mean values of different samples are shown as circles.}
\label{fig:inner_loss_wikitext103}
\end{figure*}

We perform experiments over all models on the two datasets. We attempt to study the behavior of the forward process following the same process as in Section \ref{sec:experiment_sythetic} by a cross-entropy loss with respect to the token representation of each layer by using its target label according to \eqref{eqt:inner_loss}. In this experiment, we only study the fifth or later position in a textual sequence to have enough context for the prediction task, and due to large variances, we filter out loss sequences with a final value greater than 10.0 to only focus on those cases in which the models are able to make a good prediction. Refer to Table~\ref{table:fitered_data_statistics} in the Appendix for filtered data statistics. In Figure~\ref{fig:inner_loss_wikitext103}, we observe a decreasing trend of the inner loss on all models across all of the datasets, consistent with our results in Section~\ref{sec:experiment_sythetic}. See more experimental results on the 1BW dataset in the Appendix~\ref{fig:inner_loss_1BW}. A previous work~\citep{KNN-LLM} has demonstrated that on real language data and compared to the regular forward pass, pre-trained Transformer-based CLM models achieve competitive or even better next-token prediction performances via using KNN algorithm based on intermediate representations of the models. This reflects good clustering performances of their hidden states.   

\section{A Meta-Learning View of Transformer} \label{sec:Transformer_meta}

So far, we have identified an inner optimization process within the forward process of a Transformer-based CLM model. Here, we present our meta-learning perspective built upon this inner optimization process. 

We generally view the meta-learning process as a “learning to learn” process from a variety of tasks. The “learning to learn” process is usually formulated as a bi-level optimization process. We have discovered an optimization process inside the Transformer's forward process, so training a Transformer model itself essentially forms a bi-level optimization process. The language data used for the CLM task typically spans a broad range of topics, inherently encompassing various tasks. It has long been recognized that a language model trained on such data can proficiently undertake multiple tasks~\citep{GPT-2}. 

As shown in Figures \ref{fig:learning_dynamics_f1_val} and \ref{fig:bi-level_optimization_SGD_figure}, the clustering performance and the inner loss corresponding to the inner process continue to improve across layers and training steps while the CLM loss corresponding to the outer process is decreasing over the training process. This confirms a bi-level optimization process, in which the CLM objective of the outer process can be viewed as a meta-objective. From our perspective, each Transformer layer takes context tokens as the input data and uses them to form the weightings to approximate the gradients, as discussed in Section~\ref{sec:Transformer_gc}. This can be considered as a way of learning by approximating the learning signals or gradients from close points, different from the classical stochastic gradient descent (SGD), where the gradients are explicitly computed from a loss. The learning signal comes from data in history instead of explicit training data. Based on our analysis in previous sections, the learning signal here not only contains information for minimizing the CLM loss but also of updating directions for cluster formation. Then, a Transformer layer can work as a parameterized meta-optimizer to learn to arrange the points and compute the weightings from current data to reflect expected closeness for efficiently solving a certain task, which is achieved by optimizing the meta-objective. Particularly, these close points could potentially come from different tasks such that ongoing learning could leverage knowledge from other related tasks. This conjecture is also supported by our experiments in Section \ref{sec:experiment_sythetic}, where examples generated by the same seed receive high weights such that they form a seed-specific cluster. This may indicate that the model leverages common knowledge from related examples to its input to make inferences. We leave further investigation of this property as future work. Then, the approximate gradients are used to update the current token representation to be a better one for clustering and minimizing the CLM loss. A token representation vector being continuously updated has the role of model parameters from an optimization perspective. In addition, the token representation is initialized by the current token through an embedding layer, so it also contains information about the current input. Through a series of optimization steps performed by layers, the CLM loss is expected to decrease over layers as shown in Figures \ref{fig:inner_loss_SGD} and \ref{fig:inner_loss_wikitext103}. 

Besides, by considering the next token prediction as a meta-objective, iterative clustering within the Transformer model benefits the meta-learning process by helping achieve the meta-objective. This understanding follows our intuitions in Section ~\ref{sec:language_head_knn} and is supported by the empirical result in Figures ~\ref{fig:learning_dynamics_f1_val} and ~\ref{fig:train_loss_sgd_train}, where the clustering and an inner loss computed based on a ground truth of the meta-objective are optimized simultaneously. 
    
Classical meta-learning algorithms usually optimize a large number of model parameters within their inner optimization processes such as all of the parameters of a multi-layer CNN model \citep{MAML,metaLSTM,meta_curvature}, which hinders them from efficiently using complicated optimization approaches such as higher-order optimization due to high computational cost. By contrast, we notice the number of model parameters or the number of dimensions of the token representation vector optimized by the Transformer model is usually only thousands and much smaller, potentially making the inner optimization highly efficient. Given an unseen instance, the Transformer model adjusts the weightings or the attention over the historical training data based on a potentially unseen context to produce a learning signal in each layer. This mechanism progressively reduces the inner loss for better-making predictions for the unseen instance, which is supported by results in Figure ~\ref{fig:bi-level_optimization_SGD_figure_val} in the Appendix. The process of predicting for an unseen instance is also similar to a meta-testing scenario, in which the data is split into a support set for training and a query set for testing. However, a Transformer layer---the meta-optimizer in our understanding---takes the query input (the current token) and the query context (the preceding tokens) into account for the inner process based on the support set (the historical data), which are usually too expensive to include in classical meta-learning algorithms and shows its adaptability and efficiency from another perspective. Noteworthily, the context can be imagined as a prompt to trigger a task and the task-awareness has been shown to help the meta-learning ~\citep{trident}. 

\section{Optimization Trajectory}\label{sec:optimization_energy}
In Section~\ref{sec:Transformer_gc}, we developed an understanding of the forward pass of a Transformer-based CLM model as an inner optimization process. One further question to ask is whether there are any characteristics of this optimization process. The Transformer model optimizes token representations internally. An optimization trajectory can be characterized by the evolution of a token representation. Since L2-norm regularization is commonly used with SGD as a part of an optimization process and since the token representation is the subject of optimization, we are inspired to study the property of its norm like that of model parameters to give a more comprehensive view of the inner optimization process. 

From the visualizations and quantitative analysis of the intermediate token representations in Section \ref{exp:studies_token_representation}, surprisingly, we find the norm of a token's vector representation approximately follows an incremental trend across layers in sequence. Notice the loss value has a decreasing trend at the same time across layers in sequence based on results from Sections \ref{sec:experiment_sythetic} and \ref{sec:experiment_realistic}. This is interesting because it indicates the inner optimization process may follow a specific type of optimization trajectory in general instead of an arbitrary one. In this section, we perform some analysis regarding this characteristic.  We propose the following conjecture.

\begin{conjecture}\label{hyp:norm}
A Transformer model trained for the CLM task learns a solution such that the norm of the current token vector representation is approximately non-decreasing with increasing layer number ${l}$ in the Transformer model.  
\end{conjecture} Note that we find this conjecture may not apply to the last layer of LLMs, which will be a topic for future investigation.
We also study this characteristic from a vector transformation perspective in Appendix~\ref{appx:norm_linear_model}.

\subsection{Experiments on Optimization Trajectory}\label{exp:studies_token_representation}
\begin{figure*}
\centering
\begin{subfigure}{.33\textwidth}\label{fig:visualization_GPT2_wikitext_103}
  \centering
  \includegraphics[width=0.8\linewidth]{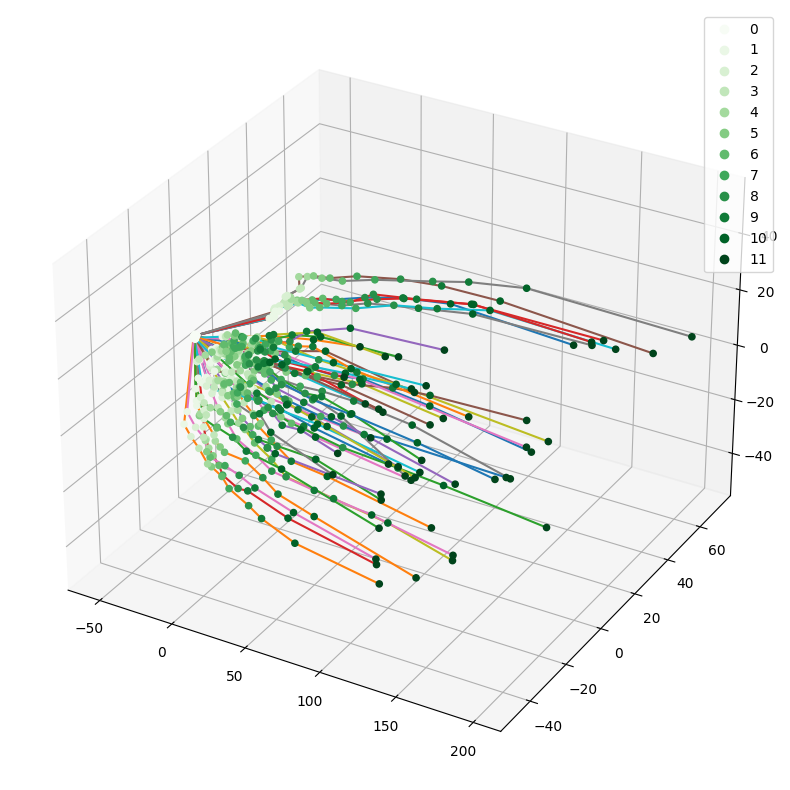}
  \caption{GPT-2}
\end{subfigure}%
\begin{subfigure}{.33\textwidth}\label{fig:visualization_llama7B_wikitext_10}
  \centering
  \includegraphics[width=0.8\linewidth]{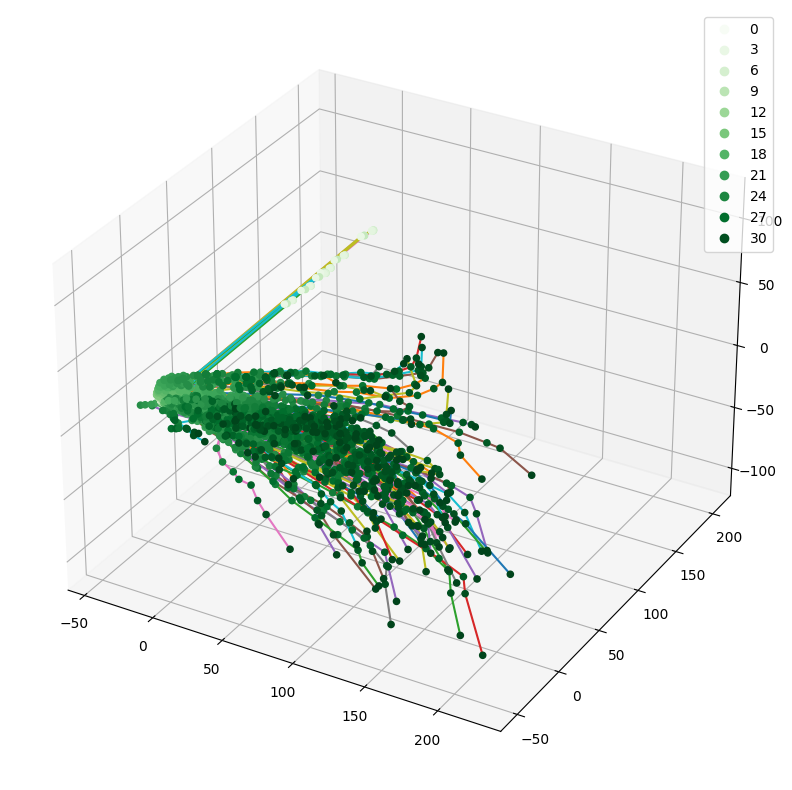}
  \caption{LLaMa-7B}
\end{subfigure}
\begin{subfigure}{.33\textwidth}\label{fig:visualization_llama13B_wikitext_10}
  \centering
  \includegraphics[width=0.8\linewidth]{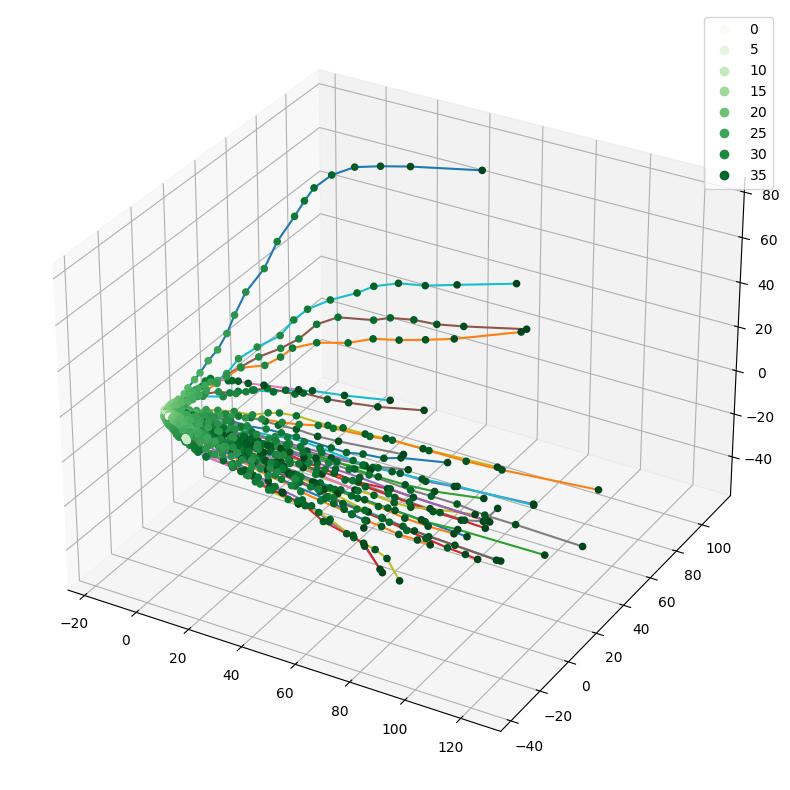}
  \caption{LLaMa-13B}
\end{subfigure}
\caption{Visuzalizations of current token vector representations for different models on the Wikitext-103 dataset.}
\label{fig:visualization_wikitext_103}
\end{figure*}

We conduct experiments based on the experimental setting in Section~\ref{sec:experiment_realistic}. We first investigate the inner optimization trajectory by visually studying how the current token vector representation evolves during the inner optimization process. We visualize the current token vector representations across layers in sequence for different samples in a 3D space by using principal component analysis (PCA)  for dimension reduction. Note that we do not consider the last layer here for better regularities because we it is not part of Conjecture~\ref{hyp:norm}. We display the optimization paths of different samples using different colors by connecting the token representations from neighboring layers. We randomly selected 50 samples for GPT-2 and 150 samples for LLaMa-7B and LLaMa-13B to make the visualizations. 

Surprisingly, we observe almost all of the samples follow a specific type of optimization trajectory instead of arbitrary ones. Table~\ref{tab:norm_percentages} shows that for the majority of samples, the norms of their current token representations are non-decreasing along the optimization trajectory and this characteristic is also not hard to see in the visualizations in Figure~\ref{fig:visualization_wikitext_103}. To study this characteristic, we measure the percentage of the optimization trajectories with non-decreasing norms (sequence-level) and the percentage of pairs of token representations at neighboring layers in a sequence whose norms are non-decreasing (pair-level). Let us assume the decreasing and non-decreasing happen uniformly at random: then the probability for a sequence of $n$ layers to have such non-decreasing property is $2^{-n}$. Considering ${n=11}$ for GPT-2, ${n=31}$ for LLaMa-7B, and ${n=39}$ for LLaMa-13B, the chance is exceedingly small. Across different models and datasets, we consistently observe a high percentage at the sequence level compared to the null random ensemble. Moreover, the pair-level percentage is almost perfect. This gives strong evidence for our Conjecture~\ref{hyp:norm}. We present additional experimental results on the 1BW dataset in the the Appendix. 

\begin{table*}[h]
\begin{center}
\begin{tabular}{lllll}
\hline
\multicolumn{1}{c}{\bf Model}  &\multicolumn{2}{c}{\bf Wikitext-103}  &\multicolumn{2}{c}{\bf 1BW}
\\ 
 & Sequence (\%) & Pair (\%) & Sequence (\%) & Pair (\%) \\
\hline
GPT-2 & 92.4 & 99.3 & 83.2 & 98.4 \\

LLaMa-7B  & 86.0 & 98.2 & 81.0 & 98.7 \\

LLaMa-13B & 79.8 & 99.0 & 73.4 & 99.0 \\
\hline
\end{tabular}
\caption{Sequence-level and pair-level measurements of the non-decreasing norm characteristic: sequence means sequence-level measurement and pair refers to pair-level measurement.}
\label{tab:norm_percentages}
\end{center}
\end{table*}

\begin{figure}[!htb]
\centering
\includegraphics[width=0.5\linewidth]{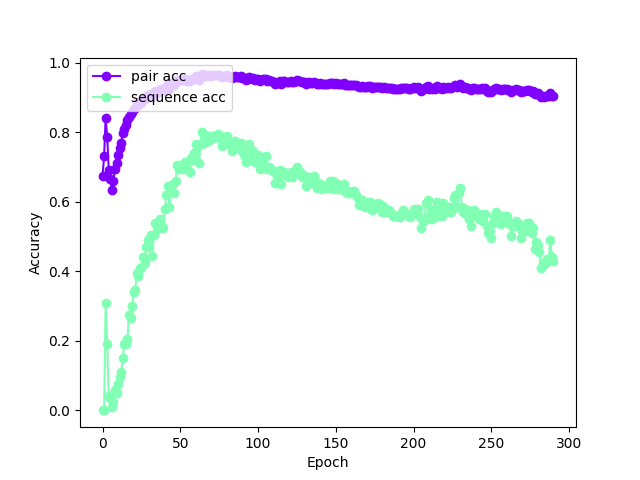}
\caption{Dynamics of both the pair-level and sequence-level accuracies along the training process based on a validation dataset. The legend shows pair-level accuracy (pair acc) and sequence-level accuracy (sequence ACC). The experimental result is obtained based on synthetic language data.}
\label{fig:norm_AdamW_var50}
\end{figure}

We also perform a similar analysis on the norm under our synthetic setting. We follow the same experimental setting in Section \ref{sec:experiment_sythetic} with the exception that we utilize a more effective AdamW optimizer~\citep{adamW} with the weight decay and a cosine annealing schedule for the learning rate to make the model converge faster, and a larger dataset to better simulate the realistic setting. More details about the new model and new data used here can be found in Appendices ~\ref{sec:appendix:synthetic_data_construction} and ~\ref{sec:appendix:hyperparameters}. We can find from Figure~\ref{fig:norm_AdamW_var50} that both the pair-level accuracy and the sequence-level accuracy improve during the early learning stage and achieve values much larger than those of the null random ensemble based on six layers. It is noticeable that the last layer of a small-scale model under our synthetic setting follows our Conjecture~\ref{hyp:norm}. 

\section{Related Work}
\subsection{Meta-Learning}
Meta-learning aims to acquire the skill of learning itself by tapping into shared knowledge among different tasks, typically through a process involving bi-level optimization. Many meta-learning methods have shown their great efficiency in learning from few-shot examples \citep{metaLSTM, MAML, FranceschiFSGP2018}. Our work focuses on studying Transformer-based CLMs from a meta-learning perspective instead of developing a new meta-learning method. \citet{Transformer_INR} treat the Transformer model as a meta-learner to learn parameters for another model but does not identify a possible inner optimization process undergone by the forward pass of the Transformer model.

\subsection{Mechanistic Interpretability}
Mechanistic interpretability involves analyzing trained neural network weights to reverse engineer the algorithms learned by the model \citep{meng2022locating, elhage2021mathematical, nanda2023emergent, ilharco2022editing, conneau2018you}. Our research also endeavors to explore hidden mechanisms of Transformer models, albeit from a distinct perspective than existing studies in this field.

\subsection{Representations of Transformer Models}
Many works have attempted to study various properties of internal representations learned by Transformer models~\citep{thompson2020topic, chen2021probing,reif2019visualizing}. Even though some works \citep{fayyaz2021not,tenney2019bert,niu2022does} have shown increasing prediction power across Transformer layers, especially following a BERT architecture \citep{BERT}, our work attempts to provide reasons for this phenomenon instead of simply presenting the existence of it. Our work also studies a special characteristic of norms of the token representations learned by the Transformer-based CLM model, which has not been explored by existing works. 

\section{Conclusion}
In this work, we show the plausibility of an inner optimization process in a Transformer model trained for the CLM task by both mathematical derivations and empirical evidence. We found the inner optimization has two-fold objectives: clustering and minimizing a CLM loss, which are complementary to each other. We also provide a detailed meta-learning view of the Transformer model trained for the CLM task, which may provide useful insights into the learning dynamics of Transformer-based causal language models. In addition, from our experiments on the inner optimization process, we discover a special characteristic about norms of the token representations learned by the Transformer-based CLM model and investigate this characteristic by theoretical analysis and experiments on real-world large language models. Our overarching objective in this study is to provide a new perspective for studying Transformer-based CLM models as well as some empirical evidence to encourage further research in this direction. The perspective may also give more insights into the design and training of more advanced Transformer models by having a clearer view of their internal mechanism.

\section{Limitations}
The main focus of this work is enhancing the understanding of the hidden mechanisms of Transformer-based CLMs. More effort can be made by inspiration from these understandings to improve existing models and propose more advanced algorithms. Our current study verifies our discovered mechanisms of Transformer-based CLMs on limited models and datasets. It is interesting to study these mechanisms in more diverse scenarios.  

\subsubsection*{Acknowledgments}
We are grateful to Sourya Basu, Akhil Bhimaraju, Moulik Choraria, Austin Ellis-Mohr, Anuj K.\ Nayak and Anay Pattanaik for their feedback on the manuscript.

\bibliography{main}
\bibliographystyle{apalike}
\appendix

\section{Derivations}
\subsection{Derivations for the MHSA module}\label{appx:MHSA}
We assume the dimension of a token representation within a Transformer model to be ${d_{model}}$. By following \citet{Transformer}, we write and expand the formulation of the multi-head attention module for the CLM task with ${z^{l} \in \mathbb{R}^{n\times d_{model}}}$ being the representations of current ${n}$ sequential tokens at layer ${l}$: 
\begin{equation}\label{eqt:MHSA}
    \begin{aligned}
       \text{MHSA}^{l}(z_{1:n}^{l}) = \sum_{h=1}^{H}W_{O}^{lh}\sum_{i=1}^{n}v_{i}^{lh}\text{softmax}((k^{lh})^\mathsf{T}q_{n}^{lh})_{i}
    \end{aligned}
\end{equation}
where ${\text{MHSA}^{l}}$ is the parameterized multi-head self-attention module at layer ${l}$. Here, ${W_{O}^{lh}} \in \mathbb{R}^{d_{model}\times d_{head}}$, ${W_{V}^{lh}}$, ${W_{K}^{lh}}$, and ${W_{Q}^{lh} \in \mathbb{R}^{d_{head}\times d_{model}}}$ are output, value, key, and query projection matrices of head ${h}$ at layer ${l}$ respectively. We have ${d_{model} = Hd_{head}}$ and ${H}$ as the total number of heads. The $i$th value ${v_{i}^{lh}}$, key ${k_{i}^{lh}}$, and query ${q_{i}^{lh}}$ of head ${h}$ at layer ${l}$ are computed by ${v_{i}^{lh} = W_{V}^{lh}z_{i}^{l}}$, ${k_{i}^{lh} = W_{K}^{lh}z_{i}^{l}}$ and ${q_{i}^{lh} = W_{Q}^{lh}z_{i}^{l}}$ respectively. Further, ${k^{lh} = \left[k_{1}^{lh}, \ldots, k_{n}^{lh}\right]}$ is the key matrix of head ${h}$ at layer ${l}$. The $i$th element of the \text{softmax} output is denoted ${\text{softmax}((k^{lh})^\mathsf{T}q_{n}^{lh})_{i}}$. Throughout, we ignore the scalar factor within the \text{softmax} attention and any bias terms, to simplify analysis.

We expand ${W_{O}^{lh}}$ by \eqref{eqt:W_GD_approx} to get an approximation: 
\begin{equation}
\begin{aligned}
   \text{MHSA}^{l}(z_{1:n}^{l}) \approx \sum_{h=1}^{H}(- \sum_{t=1}^{T}\eta_{t}^{l}\nabla_{\hat{y}_{t}^{lh}}L\otimes \hat{v}_{t}^{lh})
   \sum_{i=1}^{n}v_{i}^{lh}\text{softmax}((k^{lh})^\mathsf{T}q_{n}^{lh})_{i}  
\label{eqt:MHSA_approx}
\end{aligned}
\end{equation}
where ${L}$ is assumed to be the CLM loss throughout and we use ${\hat{\cdot}}$ to denote something in the training history. Here, ${\hat{v}_{t}^{lh}}$ refers to a historical value vector of head ${h}$ and layer ${l}$ at step ${t}$ in the training history. Note that we omit the token positions for anything in the training history and ${t}$ should not be confused with a token position within a sequence. As before, ${\hat{y}_{t}^{lh} = W_{O,t}^{lh}\hat{v}_{t}^{lh} \in \mathbb{R}^{d_{model}}}$ and ${\eta_{t}^{lh}}$ is the learning rate of the corresponding weight matrix ${W_{O,t}^{lh}}$ of head ${h}$ and layer ${l}$ at step ${t}$. 

Furthermore, by the chain rule, we have: 
\begin{equation}
\begin{aligned}
  \nabla_{\hat{y}_{t}^{lh}}L = 
  \left(\frac{\partial \hat{z}_{t}^{l}}{\partial \hat{y}_{t}^{lh}}\right)^\mathsf{T}\nabla_{\hat{z}_{t}^{l}}L  
\label{eqt:chain}
\end{aligned}
\end{equation}
where ${\hat{z}_{t}^{l}}$ is a historical token representation of layer ${l}$ at step ${t}$. 

With \eqref{eqt:chain}, we can rewrite \eqref{eqt:MHSA_approx} as: 
\begin{equation}
\begin{aligned}
   \text{MHSA}^{l}(z_{1:n}^{l}) \approx \sum_{h=1}^{H}\left[- \sum_{t=1}^{T}(\frac{\partial \hat{z}_{t}^{l}}{\partial \hat{y}_{t}^{lh}})^\mathsf{T}\eta_{t}^{l}\nabla_{\hat{z}_{t}^{l}}L\otimes \hat{v}_{t}^{lh}\right]
   \sum_{i=1}^{n}v_{i}^{lh}\text{softmax}((k^{lh})^\mathsf{T}q_{n}^{lh})_{i} \\
   = -\sum_{t=1}^{T}\eta_{t}^{l}\sum_{h=1}^{H}w_{t}^{lh}A_{t}^{lh}\nabla_{\hat{z}_{t}^{l}}L 
\label{appx:eqt:MHSA_approx_2}
\end{aligned}
\end{equation}
where $w_{t}^{lh}$ represents a weighting factor with $w_{t}^{lh} = (\hat{v}_{t}^{lh})^\mathsf{T}\sum_{i=1}^{n}v_{i}^{lh}\text{softmax}((k^{lh})^\mathsf{T}q_{n}^{lh})_{i}$, and \\ ${A_{t}^{lh} = (\frac{\partial \hat{z}_{t}^{l}}{\partial \hat{y}_{t}^{lh}})^\mathsf{T} \in \mathbb{R}^{d_{model}\times d_{model}}}$ is a transformation matrix. 

\subsection{Derivations for the FFN module}\label{appx:FFN}
We define the \text{FFN} module as follows: 
\begin{equation}
\begin{aligned}   
    \text{FFN}^{l}(z_{n}^{l+\frac{1}{2}}) = W_{2}^{l}\phi(W_{1}^{l}z_{n}^{l+\frac{1}{2}})
\label{eqt:FFN}
\end{aligned}
\end{equation}
where ${W_{2}^{l} \in \mathbb{R}^{d_{model}\times d_{ff}}}$ and ${W_{1}^{l} \in \mathbb{R}^{d_{ff}\times d_{model}}}$ are weight matrices of the first and second layer in the \text{FFN} module at Transformer layer ${l}$ and ${\phi}$ is a nonlinear activation function.

We can expand the formulation in \eqref{eqt:FFN} via \eqref{eqt:W_GD_approx} and chain rule:
\begin{subequations}\label{eqt:FFN_approx}
    \begin{equation}
        \begin{aligned} 
            \text{FFN}^{l}(z_{n}^{l+\frac{1}{2}}) \approx -(\sum_{t=1}^{T}\eta_{t}^{l+\frac{1}{2}}\nabla_{\hat{b}_{t}^{l}}L\otimes \hat{a}_{t}^{l})a_{n}^{l}\\
            = -\left[\sum_{t=1}^{T}(\frac{\partial \hat{z}_{n}^{l+\frac{1}{2}}}{\partial \hat{b}_{t}^{l}})^\mathsf{T}\eta_{t}^{l+\frac{1}{2}}\nabla_{\hat{z}_{t}^{l+\frac{1}{2}}}L\otimes \hat{a}_{t}^{l}\right]a_{n}^{l} 
        \end{aligned} 
    \end{equation}\label{eqt:FFN_approx_1}\\
    \begin{equation}
        \begin{aligned} 
            \rightarrow z_{n}^{l+1} = Norm(z_{n}^{l+\frac{1}{2}} + \text{FFN}^{l}(z_{n}^{l+\frac{1}{2}}))\\\approx Norm(z_{n}^{l+\frac{1}{2}}  - \left[\sum_{t=1}^{T}\eta_{t}^{l+\frac{1}{2}}(\frac{\partial \hat{z}_{t}^{l+\frac{1}{2}}} {\partial \hat{b}_{t}^{l}})^\mathsf{T}\nabla_{\hat{z}_{t}^{l+\frac{1}{2}}}L\otimes \hat{a}_{t}^{l}\right]a_{n}^{l}) \\
            \end{aligned} 
    \end{equation}\label{eqt:FFN_approx_2}
    \begin{equation}
        \begin{aligned} 
            = Norm(z_{n}^{l+\frac{1}{2}} - \sum_{t=1}^{T}\eta_{t}^{l+\frac{1}{2}}w_{t}^{l+\frac{1}{2}}B_{t}^{l}\nabla_{\hat{z}_{t}^{l+\frac{1}{2}}}L) 
            \\= Norm(z_{n}^{l+\frac{1}{2}} - \Bar{B^{l}\nabla_{z_{n}^{l+\frac{1}{2}}}L})
        \end{aligned} 
    \end{equation}
\end{subequations}
Here ${a_{n}^{l} = \phi(W_{1}^{l}z_{n}^{l+\frac{1}{2}})}$ and ${L}$ is the CLM loss, ${\hat{a}_{t}^{l}}$ is a historical output of the first layer at the $l$th Transformer layer ${l}$ at step ${t}$. Further, ${\hat{b}_{t}^{l} = W_{2, t}^{l}\hat{a}_{t}^{l}} \in \mathbb{R}^{d_{model}}$ and ${\eta_{t}^{l+\frac{1}{2}}}$ is the corresponding learning rate at step ${t}$, ${w_{t}^{l+\frac{1}{2}} = (\hat{a}_{t}^{l})^\mathsf{T}a_{n}^{l}}$ is a weighting factor, ${B_{t}^{l} = (\frac{\partial \hat{z}_{t}^{l+\frac{1}{2}}}{\partial \hat{b}_{t}^{l}})^\mathsf{T} \in \mathbb{R}^{d_{model}\times d_{model}}}$ is a transformation matrix, and ${\Bar{B^{l}\nabla_{z_{n}^{l+\frac{1}{2}}}L}}$ is an approximation of the transformed gradient with respect to the current input ${z_{n}^{l+\frac{1}{2}}}$. 

\section{Study of Conjecture \ref{hyp:norm} from a vector transformation perspective} \label{appx:norm_linear_model}
Due to the complexity of the nonlinear transformation of a Transformer layer, we initiate studies on this characteristic by using a simplified linear model. We begin by simplifying the \text{softmax}-based attention mechanism as a linear attention mechanism as done in some prior works~\citep{in_context_gd,in_context_gpt} such that the \text{MHSA} module becomes:
\begin{equation}\label{eqt:MHSA_linear}
    \begin{aligned}
       & \text{MHSA}_{linear}^{l}(z^{l}_{1:n}) = \sum_{h=1}^{H}W_{O}^{lh}\sum_{i=1}^{n}W_{V}^{l}z_{i}^{l} 
 (W_{K}^{lh}z_{i}^{l})^\mathsf{T}W_{Q}^{lh}z_{n}^{l} \\
       & = \sum_{h=1}^{H}\sum_{i=1}^{n}W_{O}^{lh}W_{V}^{l}z_{i}^{l}(z_{i}^{l})^\mathsf{T}(W_{K}^{lh})^\mathsf{T}W_{Q}^{lh}z_{n}^{l} \\ 
       & = W_{\text{MHSA}}^{l}z_{n}^{l}
    \end{aligned}
\end{equation}
by following conventions from \eqref{eqt:MHSA}. We can now treat the \text{MHSA} module as a linear transformation using matrix ${W_{\text{MHSA}}^{l}}$. 

Similarly, we obtain a linear version of the \text{FFN} module by removing its nonlinear activation function:
\begin{equation}\label{eqt:FFN_linear}
    \begin{aligned}
       \text{FFN}_{linear}^{l}(z_{n}^{l+\frac{1}{2}}) = W_{2}^{l}W_{1}^{l}z_{n}^{l+\frac{1}{2}} = W_{\text{FFN}}^{l}z_{n}^{l+\frac{1}{2}}.
    \end{aligned}
\end{equation}

We develop a linear model corresponding to a Transformer layer by omitting the layer normalization and combining everything and residual connections: 
\begin{equation}\label{eqt:tf_layer_linear}
    \begin{aligned}
       z_{n}^{l+1} = (I + W_{\text{FFN}}^{l})(I + W_{\text{MHSA}}^{l})z_{n}^{l} = W_{linear}^{l}z_{n}^{l}
    \end{aligned}
\end{equation}
where ${I \in \mathbb{R}^{d_{model}\times d_{model}}}$ is an identity matrix. 

We present the following proposition for the norm of the current token representations across layers in the linear model.
\begin{prop}\label{prop:weak}
Let ${W \in \mathbb{R}^{d_{in} \times d_{out}}}$ be a linear transformation matrix, and ${x \in \mathbb{R}^{d_{in}}}$ and ${y \in \mathbb{R}^{d_{out}}}$ be the input and output of the linear transformation. We can have the Gram matrix of ${W}$ decomposed by eigendecomposition as ${W^\mathsf{T}W = U\Lambda U^{-1}}$. Let ${a = U^\mathsf{T}x}$ and ${b = U^{-1}x}$. Given ${\sum_{i=1}^{d_{out}}\lambda_{i}a_{i}b_{i} \ge \sum_{i=1}^{d_{out}}a_{i}b_{i}}$, we have $\norm{y} \ge \norm{x}$, where ${\lambda_{i}}$ is the $i$th eigenvalue of the Gram matrix. 
\end{prop}

See a proof in Appendix \ref{appx:prop2_proof}. With Proposition~\ref{prop:weak}, we make another conjecture.
\begin{conjecture}\label{hyp:eigenvalue}

One of the intrinsic characteristics of ${W_{linear}^{l}}$ in the aforementioned linear model constructed from a Transformer model trained for the CLM task is to have large enough eigenvalues for its Gram matrix such that its output has norm no less than that of its input.  

\end{conjecture}

A way to theoretically guarantee the non-decreasing property of the norm is to transform the current token representation via the linear model that we constructed, so as to meet the condition in Proposition \ref{prop:weak}. 

Based on this theoretical analysis, we construct the linear transformation matrices for GPT-2 for different layers and samples collected by following Section \ref{sec:experiment_realistic} from both Wikitext-103 and 1BW datasets. We found the eigenvalues of the constructed matrices perfectly meet the condition proposed in Proposition~\ref{prop:weak}, which means the matrix can effectively stretch the current vector representation resulting in an incremental norm. Note that we only conduct this experiment on GPT-2, because LLaMa uses a special \text{FFN} module with a gating mechanism instead of a standard one. 

\section{Proofs}
\subsection{Proof of Proposition~\ref{prop:weak}}\label{appx:prop2_proof}
\begin{proof}
Since ${W^\mathsf{T}W}$ is a symmetric matrix and has non-negative real eigenvalues,  
we can decompose ${W^\mathsf{T}W}$ by eigendecomposition:
${W^\mathsf{T}W = U\Lambda U^{-1}}$. \\
    \begin{equation}\label{eqt:proof_prop2}
        \begin{aligned}
           \norm{y} = \sqrt{y^\mathsf{T}y} = \sqrt{x^\mathsf{T}W^\mathsf{T}Wx} \\ 
        = \sqrt{x^\mathsf{T}U\Lambda U^{-1}x} = \sqrt{a^\mathsf{T}\Lambda b} \\ 
        = \sqrt{\sum_{i=1}^{d_{out}}\lambda_{i}a_{i}b_{i}} \ge \sqrt{\sum_{i=1}^{d_{out}}a_{i}b_{i}} \\ 
        = \sqrt{x^\mathsf{T}U U^{-1}x} = \sqrt{x^\mathsf{T}Ix} \\
        = \sqrt{x^\mathsf{T}x} = \norm{x}
        \end{aligned}
    \end{equation}
\end{proof}

\section{Synthetic Language Data Construction}
\label{sec:appendix:synthetic_data_construction}

To generate synthetic language data, we first sample a number of regular expressions as seeds. Regular expressions represented by character sequences, establish patterns for searching. They find extensive applications in tasks like text processing, string manipulation, and pattern matching. These expressions incorporate ordinary characters (such as letters and digits) alongside special characters, also termed metacharacters, which possess distinct meanings. Metacharacters empower users to define rules and criteria for identifying patterns within text. We use a regular expression in a reverse manner to sample strings matching the search pattern defined by the regular expression. 

Please see Table~\ref{table:data_hyperparameters} for related hyperparameters. We simulate sampling from different distributions by sampling from different seeds defined by regular expressions. Besides, we show the statistics of the synthetic language dataset in Table~\ref{table:data_statistics}.

\begin{table*}
\centering
\begin{tabular}{lll}
\hline
\textbf{Setting}  & \textbf{Set} & \textbf{Size} \\
\hline
Synthetic small & Training & 341\\
 & Validation & 200\\
Synthetic large & Training & 2,501\\
 & Validation & 200\\
Wikitext-103 & Testing & 2,279\\
1BW & Testing & 5,000\\
\hline
\end{tabular}
\caption{\label{table:data_statistics}
Data statistics of various language datasets used in this work. The size of a data set is quantified by its number of instances. Only testing sets are available for the real-world language datasets (Wikitext-103 and 1BW) since we use pre-trained models instead of training and validating a model in this setting. Synthetic small and Synthetic large are used by experiments in Section~\ref{sec:experiment_sythetic} and in Section~\ref{exp:studies_token_representation} respectively.
}
\end{table*}

\begin{table*}
\centering
\begin{tabular}{lll}
\hline
\textbf{Model}  & \textbf{Wikitext-103} & \textbf{1BW} \\
\hline
GPT-2 & 398 & 929\\
LLaMa-7B & 1432 & 2981\\
LLaMa-13B & 1749 & 3690\\
\hline
\end{tabular}
\caption{\label{table:fitered_data_statistics}
Data statistics of data after filtering samples with loss values larger than 10.0 by various models}
\end{table*}

\section{Hyperparameters}
\label{sec:appendix:hyperparameters}
Table~\ref{table:data_hyperparameters} presents the hyperparameters of the data generation process. Tables \ref{table:model_hyperparameters_1} and \ref{table:model_hyperparameters_2} show the hyperparameters of our Transformer models and their trainings.

\begin{table*}
\centering
\begin{tabular}{ll}
\hline
\textbf{Hyperparameter} & \textbf{Value} \\
\hline
Number of different seeds & 10 (50) \\
Maximum number of sequences per seed & 60 \\
Minimum number of sequences per seed & 10 \\
Number of seeds in the validation set & 10\\
Number of number of sequences in the validation set & 20\\
Size of vocabulary & 28\\
\hline
\end{tabular}
\caption{\label{table:data_hyperparameters}
Hyperparameters of the language data generation process. Numbers in parentheses are used for the larger synthetic dataset in Section~\ref{exp:studies_token_representation}.
}
\end{table*}

\begin{table*}
\centering
\begin{tabular}{ll}
\hline
\textbf{Hyperparameter} & \textbf{Value} \\
\hline
Learning rate & 1.0 \\
Number of epochs & 174 \\
Optimizer & SGD \\
Max gradient norm & 1.0 \\
Number of layers & 6 \\
Number of heads & 8 \\
Hidden dimension & 64 \\
Feedforward network dimension & 128 \\
Dropout & 0.2 \\
\hline
\end{tabular}
\caption{\label{table:model_hyperparameters_1}
Hyperparameters of our model in Section \ref{sec:experiment_sythetic} and its training. 
}
\end{table*}

\begin{table*}
\centering
\begin{tabular}{ll}
\hline
\textbf{Hyperparameter} & \textbf{Value} \\
\hline
Learning rate & 1E-3 \\
Number of epochs & 300 \\
Optimizer & AdamW \\
Weight decay & 0.1 \\
Max gradient norm & 1.0 \\
Scheduler & Cosine Annealing \\
Number of layers & 6 \\
Number of heads & 8 \\
Hidden dimension & 64 \\
Feedforward network dimension & 128 \\
Dropout & 0.2 \\
\hline
\end{tabular}
\caption{\label{table:model_hyperparameters_2}
Hyperparameters of our model in Section \ref{exp:studies_token_representation} and its training.
}
\end{table*}

\begin{table*}
\centering
\begin{tabular}{lll}
\hline
\textbf{Model} & \textbf{Hidden Dimension} & \textbf{Parameter Count}\\
\hline
Our model & 64 &  204 thousand\\
GPT-2  & 768 &  117 million\\
LLaMa-7B & 4096 & 7 billion \\
LLaMa-13B & 5120 & 13 billion \\
\hline
\end{tabular}

\caption{\label{table:model_size}
Sizes of models measured by parameter counts and size of hidden dimension.
}
\end{table*}

\section{Additional Experimental Results}

We show additional experimental results here. 
\begin{figure*}[!htb]
\centering
\begin{subfigure}{.3\textwidth}\label{fig:loss_GPT2_1BW}
  \centering
  \includegraphics[width=0.75\linewidth]{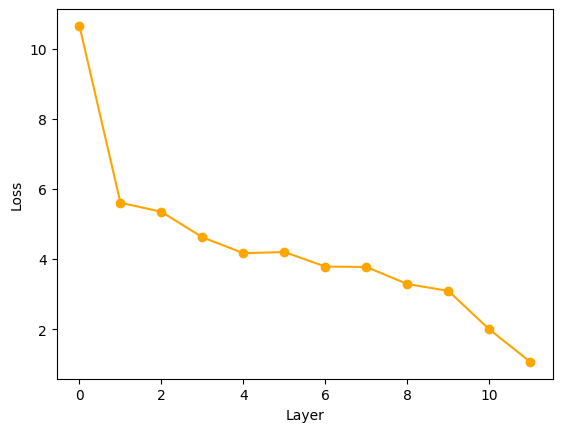}
  \caption{GPT-2}
\end{subfigure}%
\begin{subfigure}{.3\textwidth}\label{fig:loss_llama_7B_1BW}
  \centering
  \includegraphics[width=0.75\linewidth]{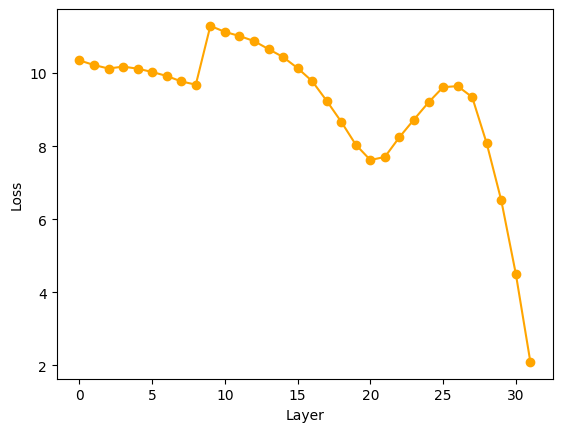}
  \caption{Llama-7B}
\end{subfigure}
\begin{subfigure}{.3\textwidth}\label{fig:loss_llama_13B_1BW}
  \centering
  \includegraphics[width=0.75\linewidth]{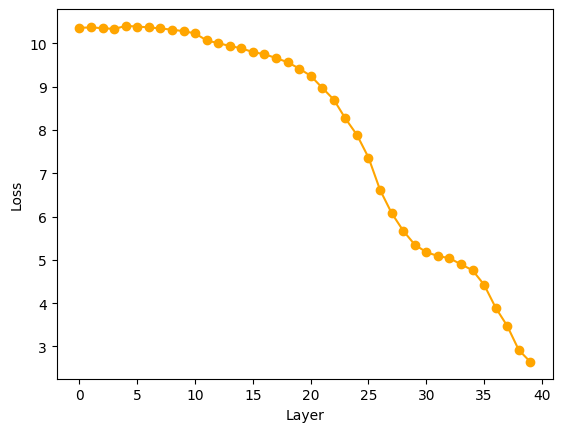}
  \caption{Llama-13B}
\end{subfigure}
\caption{Inner optimization loss across layers based on The 1BW Dataset: The x-axis is layer numbers
and y-axis is the inner loss averaged over samples. The mean values of different samples are shown as circles.}
\label{fig:inner_loss_1BW}
\end{figure*}

\begin{figure*}
\centering
\begin{subfigure}{.33\textwidth}\label{fig:visualization_GPT2_1BW}
  \centering
  \includegraphics[width=0.8\linewidth]{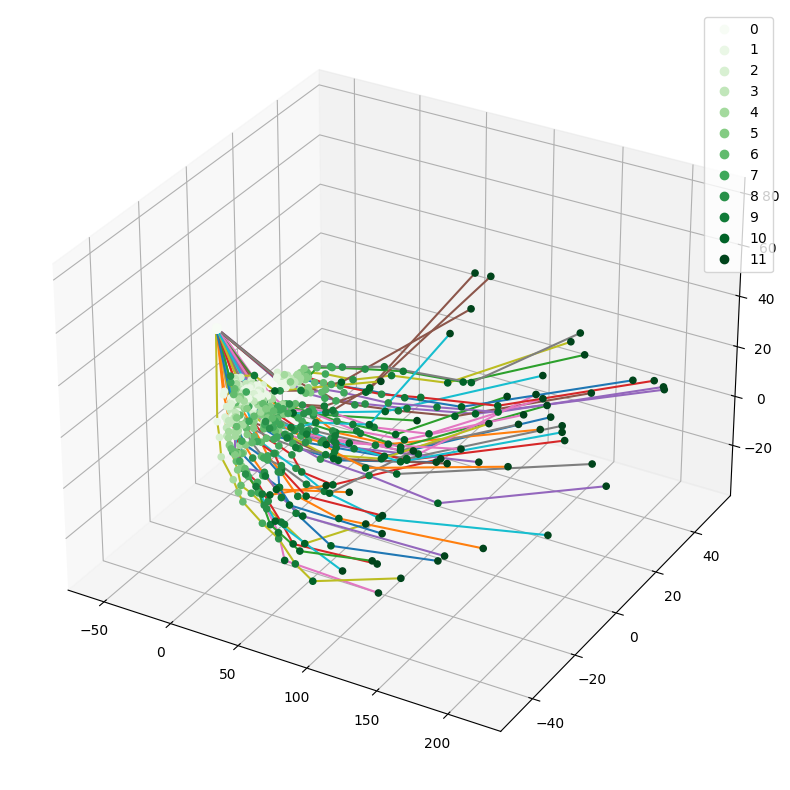}
  \caption{GPT-2}
\end{subfigure}%
\begin{subfigure}{.33\textwidth}\label{fig:visualization_llama7B_1BW}
  \centering
  \includegraphics[width=0.8\linewidth]{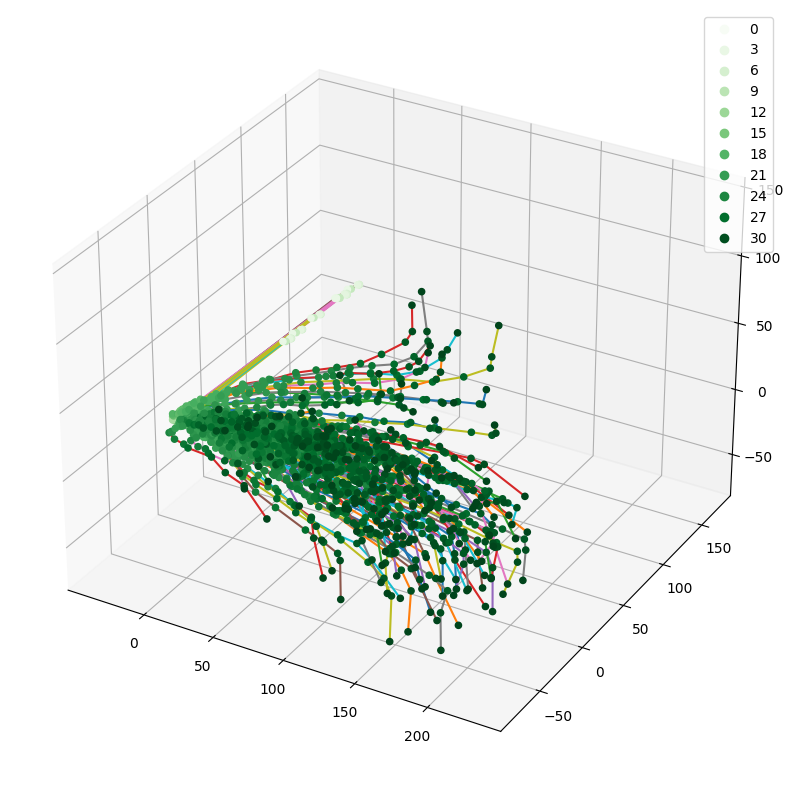}
  \caption{LLaMa-7B}
\end{subfigure}
\begin{subfigure}{.33\textwidth}\label{fig:visualization_llama13B_1BW}
  \centering
  \includegraphics[width=0.8\linewidth]{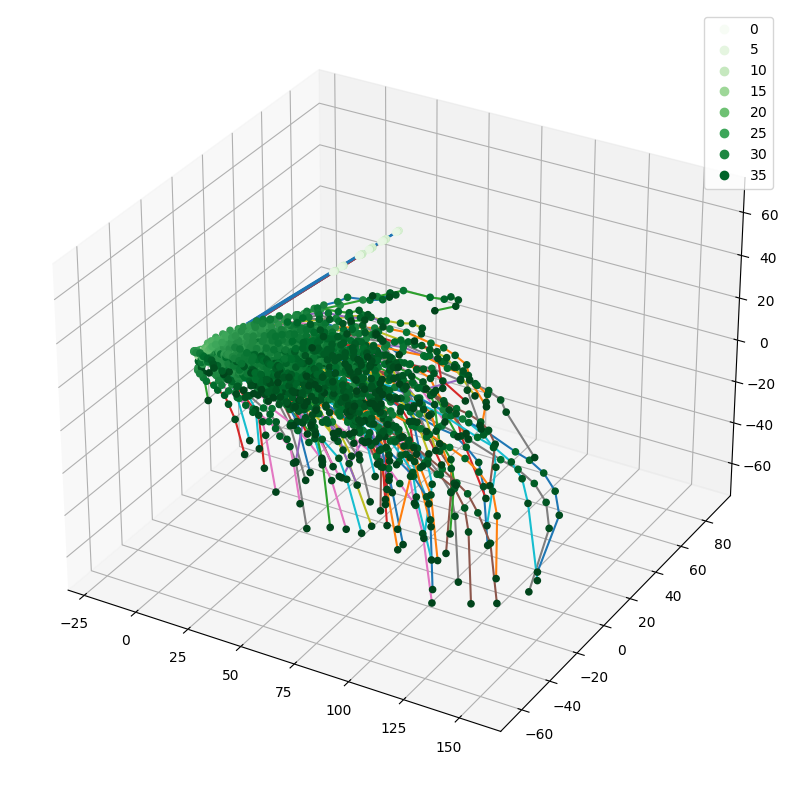}
  \caption{LLaMa-13B}
\end{subfigure}
\caption{Visuzalizations of current token vector representations for different models on the 1BW Dataset.}
\label{fig:visualization_1BW}
\end{figure*}

\label{sec:appendix:more_results}

\begin{figure*}[!htb]
\centering
\begin{subfigure}{\textwidth}\label{fig:syn_all_trainsub}
  \centering
  \includegraphics[width=0.7\linewidth]{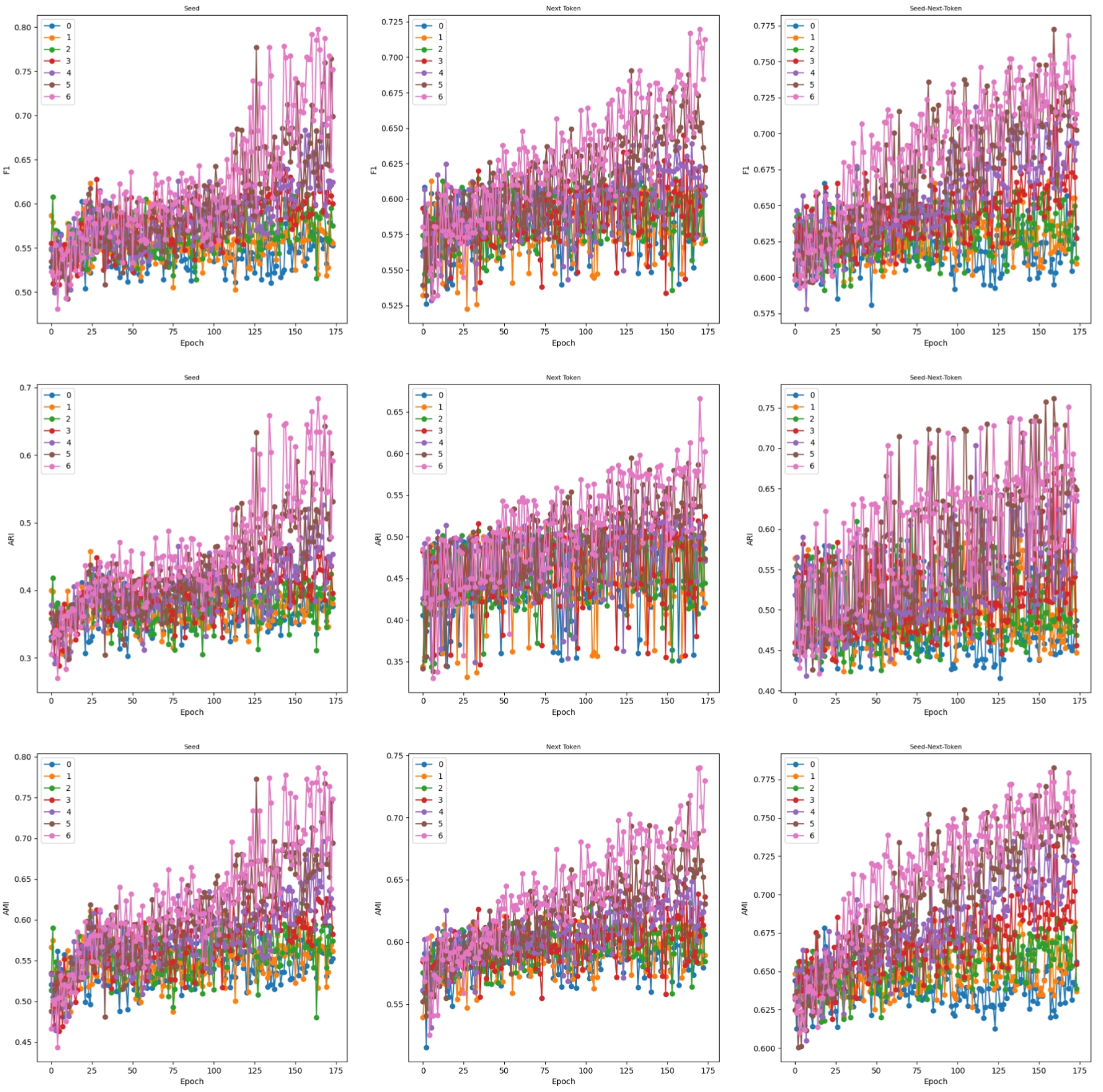}
  \caption{Training set}
\end{subfigure}%
\\
\begin{subfigure}{\textwidth}\label{fig:syn_all_val}
  \centering
  \includegraphics[width=0.7\linewidth]{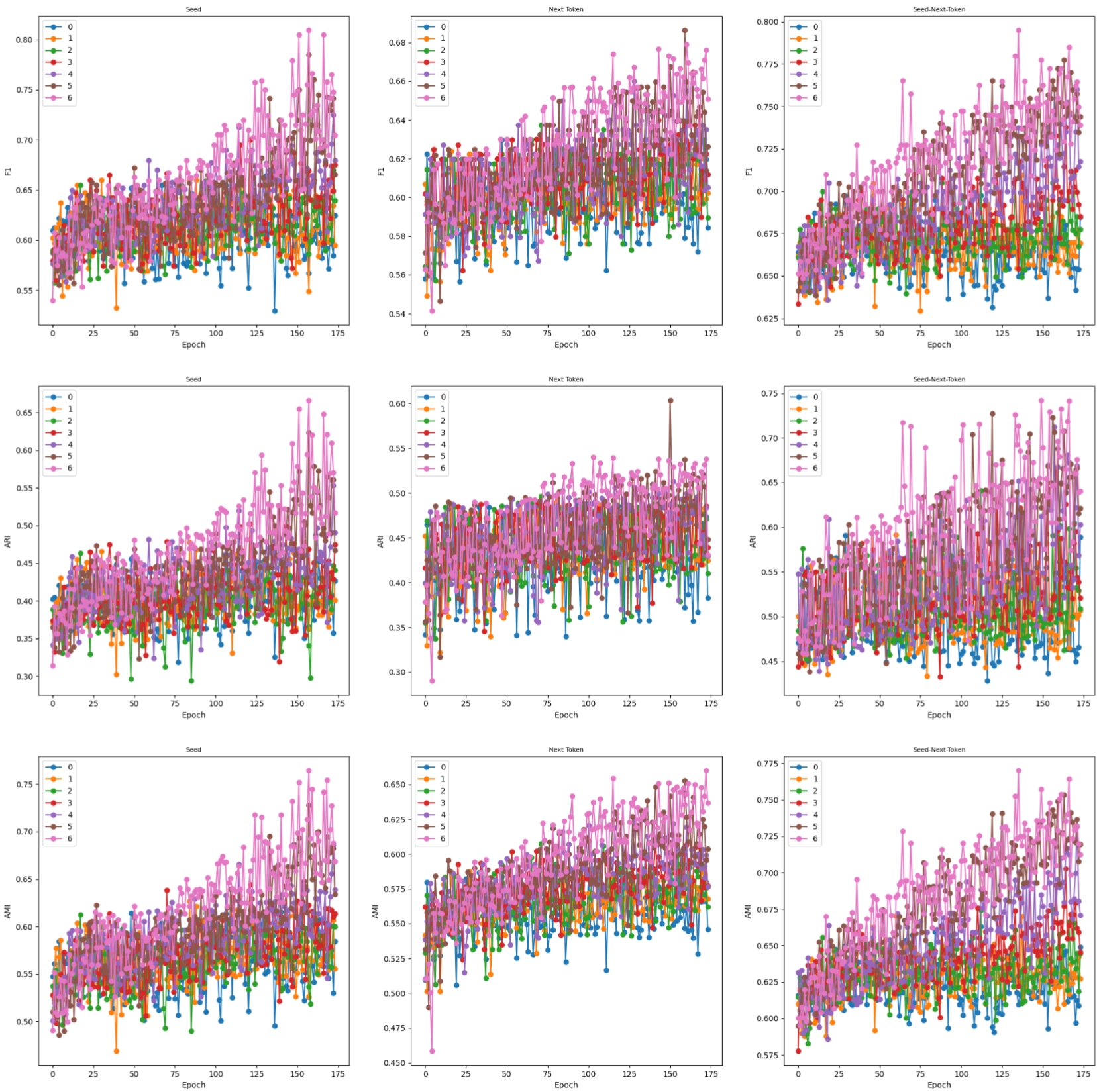}
  \caption{Validation set}
\end{subfigure}%
\caption{Clustering analysis on both the training set and the validation set across different layers throughout the training process. Different
columns indicate different ground truth labels: seed, next token, and their combination (Seed-Next-Token). The
legend shows layers. Each dot illustrates a data point.}
\label{fig:learning_dynamics_all}
\end{figure*}

\begin{figure*}[!htb]
\centering
\includegraphics[width=\linewidth]{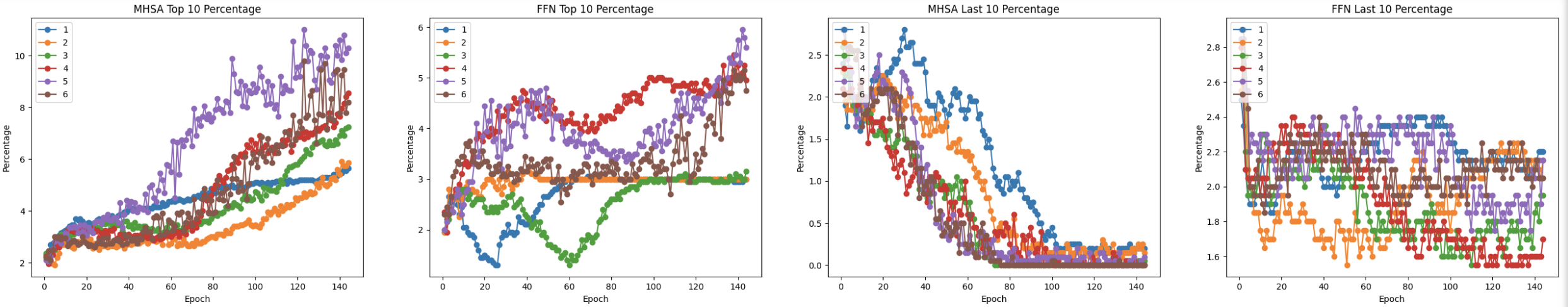}
\caption{Attention over the history throughout the training evaluated using a combination of seed and next token as ground truth: we compute the attentions from an unseen instance (the last token of each sequence) from the validation set to all of the historical instances and calculate the percentage of the top 10  and last 10 attended instances having the same label. The final measurement is aggregated across different heads, token positions, and instances and is reported per layer.}
\label{fig:att_lout=1}
\end{figure*}

\begin{figure*}[!htb]
\centering
\begin{subfigure}{0.45\textwidth}
  \centering
  \includegraphics[width=0.8\linewidth]{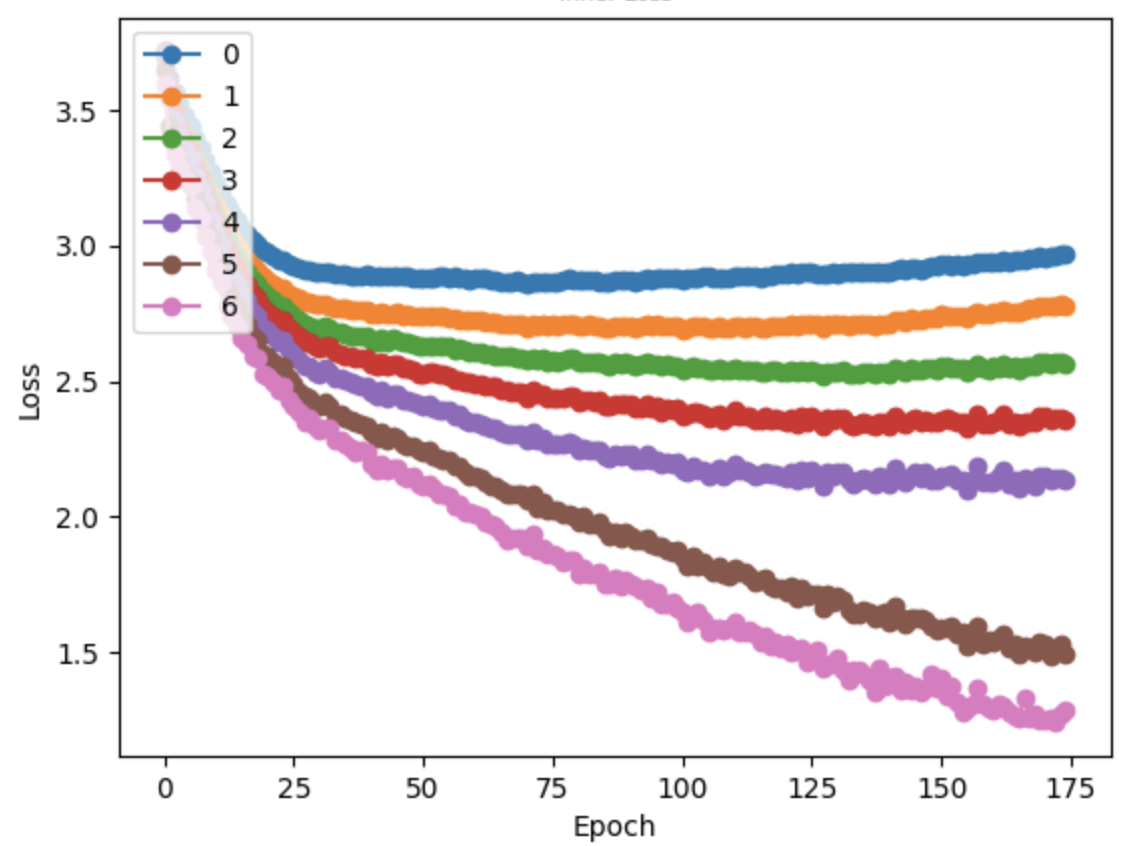}
  \caption{Inner Loss}
  \label{fig:inner_loss_SGD_val}
\end{subfigure}%
\begin{subfigure}{0.5\textwidth}
  \centering
  \includegraphics[width=\linewidth]{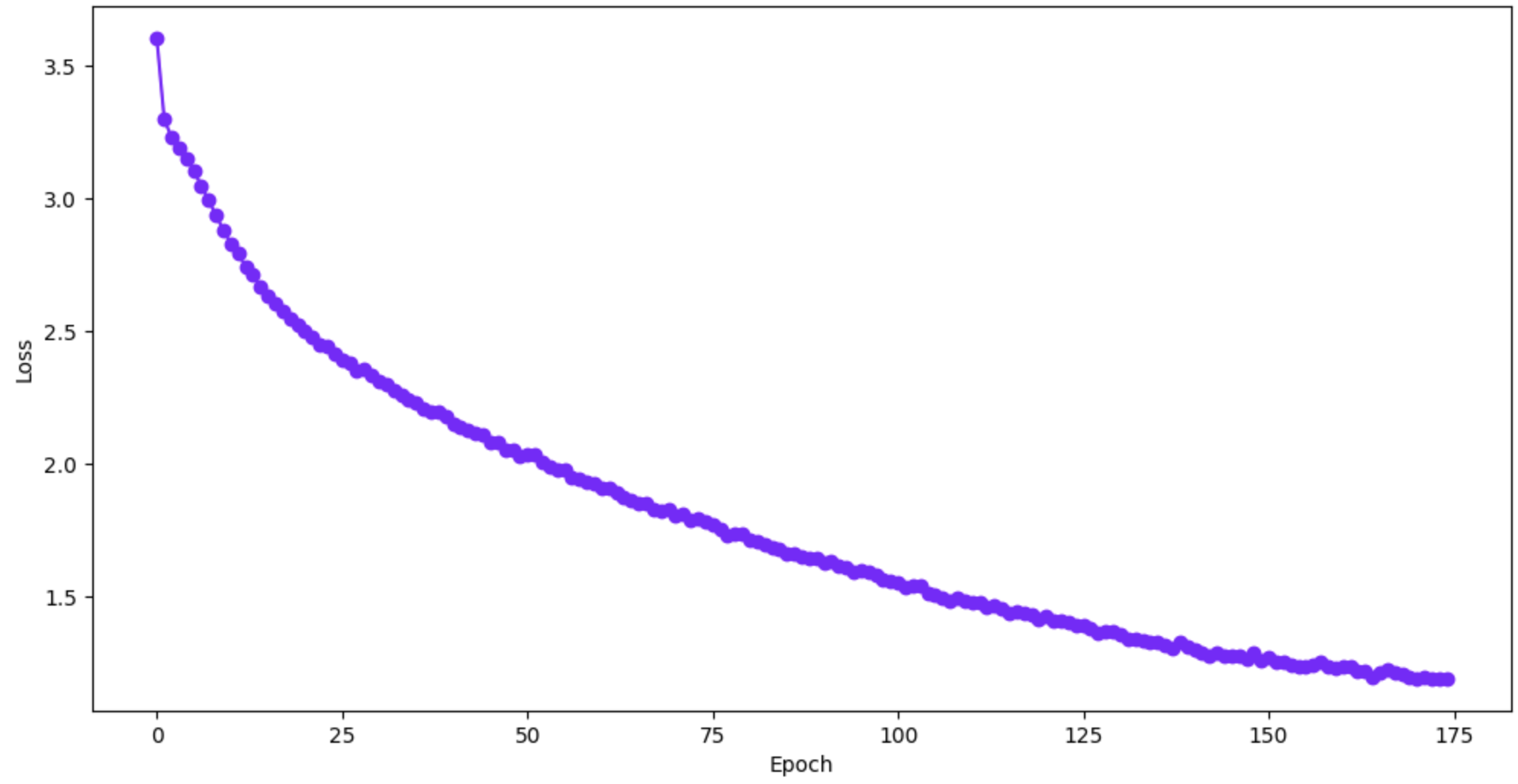}
  \caption{Outer Loss (Validation Loss)}
  \label{fig:val_loss_sgd_val}
\end{subfigure}%
\caption{Bi-level Optimization process within a Transformer-based CLM model. (a) shows inner optimization losses across layers and training epochs computed according to \eqref{inner_loss_equation} and aggregated from validation examples. (b) illustrates the losses of the outer optimization process throughout the training process, which explicitly optimizes for the CLM task. The outer loss is identical to the validation loss of the model.}
\label{fig:bi-level_optimization_SGD_figure_val}
\end{figure*}

\end{document}